\documentclass[minf,frontabs,twoside,singlespacing,parskip,deptreport]{infthesis}  
\usepackage{graphicx}
\usepackage{pdfpages}
\usepackage{subcaption}
\usepackage{hyperref}
\usepackage{multirow}
\usepackage[figuresleft]{rotating}
\graphicspath{{./images/}}
\begin{document}

\title{Extractive text summarisation of Privacy Policy documents using machine learning approaches}

\author{Chanwoo Choi}

\course{Artificial Intelligence and Computer Science}

\project{5th Year Project Report}

\date{2022}

\abstract{
 This work demonstrates two Privacy Policy (PP) summarisation models based on two different clustering algorithms: K-means clustering and Pre-determined Centroid (PDC) clustering. K-means is decided to be used for the first model after an extensive evaluation of ten commonly used clustering algorithms. The summariser model based on the PDC-clustering algorithm summarises PP documents by segregating individual sentences by Euclidean distance from each sentence to the pre-defined cluster centres. The cluster centres are defined according to General Data Protection Regulation (GDPR)'s 14 essential topics that must be included in any privacy notices. The PDC model outperformed the K-means model for two evaluation methods, Sum of Squared Distance (SSD) and ROUGE by some margin (27\% and 24\% respectively). This result contrasts the K-means model's better performance in the general clustering of sentence vectors before running the task-specific evaluation. This indicates the effectiveness of operating task-specific fine-tuning measures on unsupervised machine-learning models. The summarisation mechanisms implemented in this paper demonstrates an idea of how to efficiently extract essential sentences that should be included in any PP documents. The summariser models could be further developed to an application that tests the GDPR-compliance (or any data privacy legislation) of PP documents.
 
 \textbf{Keywords:} Privacy policy, clustering, text summarisation, pre-determined centroids
}

\maketitle

\section*{Acknowledgements}
I would like to thank my project supervisor Dr Robin Hill for his descriptive feedback and his efforts in providing helpful comments during semi-weekly meetings. I would also like to acknowledge Dr Nadin Kokciyan for providing comments and ideas for the project. I also want to thank my parents for all their financial and emotional support.
\tableofcontents


\chapter{Introduction}

When registering to a website, users are explicitly required to click the "I read and agree" button for the privacy notice of the website to proceed with the registration process and be able to use the product. If a user does not agree to the privacy policy as he was not sure about certain clauses, the user simply cannot access the service. Therefore, most people mindlessly click the button and move on, according to a study performed by Obar et al. \cite{obar2020biggest}. In the study, it was observed that about three-quarters of the participants agreed to the privacy notice without reading the document at all. The remaining quarter of the users spent less than two minutes on average on a document that requires a 20-minute reading time for proper comprehension. Users are easily overwhelmed by the huge amount of information written in legalese and the user-unfriendly design of the document \cite{steinfeld2016agree}. Website owners often produce Privacy Policy (PP) documents with poor readability and accessibility, and the authors even suggest that this might be an intentional behaviour to demotivate users from reading certain sensitive contents.

However, it is rather unreasonable to expect all users to read the documents thoroughly every time they register for or use an online service. There are 1.88 billion websites on the internet as of 2021 \cite{armstrong_2021}. According to data privacy legislation such as General Data Protection Regulation (GDPR) \cite{voigt2017eu}, website owners are legally required to write and display privacy notices to their users. This implies that there should be about 1.88 billion Privacy Policy (PP) documents on the internet. If all users listened to the security advice from the government and from other sources and started reading the documents as suggested, the social cost of such behaviour would be extremely huge making it absurdly counterproductive. According to \cite{mcdonald2008cost}, the total economic value of the supposed reading time (this study is limited to the US population only) is estimated to \$781 billion assuming extensive and comprehensive reading, and \$492 billion when just skimmed through. Understanding that the number of websites on the internet (and therefore PP documents) increased drastically compared to when this study was performed in 2008, the cost of forcing people to read all the documents every time would be too inefficient in so many ways. 

There have been numerous suggestions to make reading PP documents more efficient and cheaper. \cite{cherivirala2016visualization} and \cite{liu2018towards} introduced automatic classification of sentences in PP documents. By classifying the documents, users could only spend a fraction of the time reading the parts that are more relevant and of direct interest to them. However, this method is somewhat limited considering that people are still needed to educate themselves to figure out which parts (sections) of PP documents they feel more sensitive to or find themselves to be more interested in. \cite{bannihatti2020finding} implements an opt-out choices detector that spots opt-out choices that usually include sensitive sections but are not so easy to be found \cite{habib2020s}. Typical opt-out choices include agreeing to sharing users' data with third-party bodies or using them for marketing purposes, etc. \cite{westin2019opt} reveals that websites often manipulate users' behaviours using "dark patterns" to agree to potentially risky or less secure content, speaking out to the tech community to design software applications and websites to be more humane and secure (in terms of privacy protection). As an effort to motivate users to read PP documents and make readings of them more efficient, I implemented in my previous work \cite{choi2020worth} a two-class PP sentence chunk classifier that indicates "more important" clauses based on three personas with different level of sensitivity to data privacy. Further details would be covered in section 2.1. 

As an extension to my previous work \cite{choi2020worth}, this paper implements text summarisation models for sentences in PP documents using machine-learning based clustering algorithms. The models explained in this paper do not select sentences based on their "level of risk" or "data sensitivity". The goal of the models is to extract sentences that cover essential topics following the GDPR guideline \cite{wodford_2019} so that users only need to put the minimum effort to understand the gist of the documents. The rest of the paper is structured as the following: First, there would be a chapter on the background knowledge and understanding of the technical concepts used in this work and related works, including reflections on my previous work \cite{choi2020worth}. Then, a chapter to explain the detailed methodology of how I implemented the summarisation mechanisms and justify my technical choices during implementation. The next chapter illustrates how the summarisation models are evaluated using two different summary evaluation methods: Sum of Squared Distance (SSD) and ROUGE \cite{lin2004rouge}, followed by an explanation of the results in further detail. The last chapter would conclude the work by describing the findings, limitations of this work and how would I overcome the limitations and improve them in the future.

\chapter{Background}

\section{Reflections on previous work}

As a means to facilitate the readings of Privacy Policy documents and motivate users, I introduced a text classifier in my previous work \cite{choi2020worth} that categorises each clause from a given Privacy Policy document into two classes: "worth reading" or "standard/trivial". The classifier works in 3 levels of sensitivity to data privacy, highlighting a larger number of "important" clauses to users with a higher level of sensitivity. The three personas are elaborated in Table \ref{tab:personas}. Here, it is assumed that individuals with a higher level of knowledge of how their data are processed in software applications would be less sensitive to handing over their data as they know (from the documents) what the companies are doing with their personal data. From a study conducted in \cite{choi2020worth} to evaluate the performance and the usability of the classifier, participants mentioned that they are unlikely to use the product even though its functionalities and overall performances are improved. This is probably because the users were unaware of the significance of these policies and remained unmotivated to read them at all. 

\begin{table}[!h]
\centering
\begin{tabular}{|c|c|c|}
	\hline
	Persona & Data privacy sensitivity & Age and occupation \\
	\hline\hline
	Alice & High & 55 yrs, Farmer\\
	\hline
	Bob & Medium & 45 yrs, Office worker\\
	\hline
	Charlie & Low & 25 yrs, Software engineer\\
	\hline
\end{tabular}
\caption{The three personas and their attributes.}
\label{tab:personas}
\end{table}

\begin{figure}[h!]
\includegraphics[width=\textwidth]{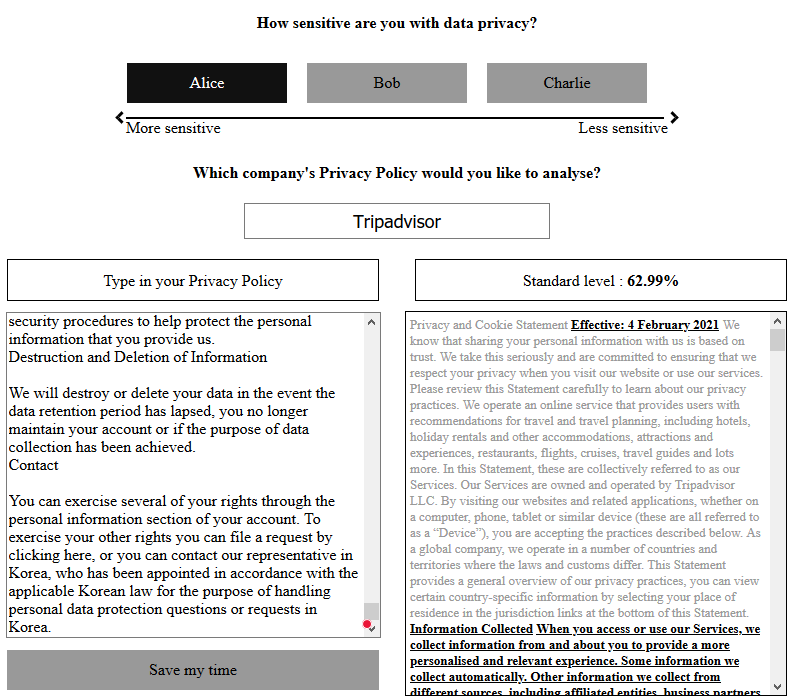}
\centering
\caption{The demonstration of "Worth reading finder" as illustrated in my last work \cite{choi2020worth}}
\label{fig:frontendAlice}
\end{figure}

 The prototype solution introduced in \cite{choi2020worth} only achieved its goal partially as it was unsuccessful in strongly motivating users to find a clear reason to read through the heavy documents. The result of the study suggests the necessity of an automated solution to motivate users even further, giving them the impression that they could read and understand the necessary part of the document only by providing a minimum effort. To achieve this, this paper suggests two PP document summarisation mechanisms whose primary aim is to only take essential sentences from the documents, based on two different clustering algorithms: K-means clustering and pre-determined Centroid (PDC) Clustering. Here, the essential clauses are chosen based on the specification suggested in "Privacy Notice best practice and Privacy Notice template" provided by GDPR \cite{wodford_2019}.

\section{Clustering}

In the modern days, clustering refers to unsupervised machine-learning algorithms that have been very effective in understanding (or at least briefly visualising) large datasets with limited knowledge. Before machine learning became so popular, the idea of clustering was first introduced in 1955 by Steinhaus, and there has been great research in the domain \cite{jain2010data}. There are numerous algorithms that implement clustering, and as explained in the previous section, this paper uses K-means clustering and pre-determined centroid clustering (PDC clustering). PDC clustering is yet to be actively researched by many scientists and it is hard to find a renowned implementation for this idea as well. However, in terms of its concept, it resembles multiple semi-supervised or supervised clustering algorithms, such as \cite{eick2004supervised}, \cite{basu2002semi}, \cite{bair2013semi} and \cite{zhu2019classification}. Both clustering algorithms use Euclidean distance to find the nearest neighbours from centroids. The difference is that K-means clustering chooses random points at the start as centroids, and PDC performs clustering based on the given set of centre points. The reason that K-means is chosen for this work among various unsupervised clustering algorithms is its effectiveness on the current dataset and task, as it achieves the highest Silhouette score compared to other algorithms. This is explained in further detail in section 3.2.3. 

\section{Transformers}

Modern text encoders and generators are often based on transformers \cite{vaswani2017attention}, a new structure of neural network for Natural Language Processing (NLP) tasks dealing with sequential text data. It is introduced to overcome the limitations of Recurrent Neural Networks (RNN) \cite{mikolov2010recurrent} and Long Short-term memory (LSTM) \cite{hochreiter1997long}, especially the heavy hardware requirements and excessive time resources needed for training the networks. Unlike RNNs and LSTMs which always have to be fed non-parallel sequential data, Transformers take multiple sequences of words in parallel, reducing the training time and requiring fewer hardware resources. Due to their robustness and relatively cost-effective nature, transformers are now actively used by leading researchers and institutions for state-of-the-art performances for certain NLP tasks.

\begin{figure}[h!]
\includegraphics[width=0.55\textwidth]{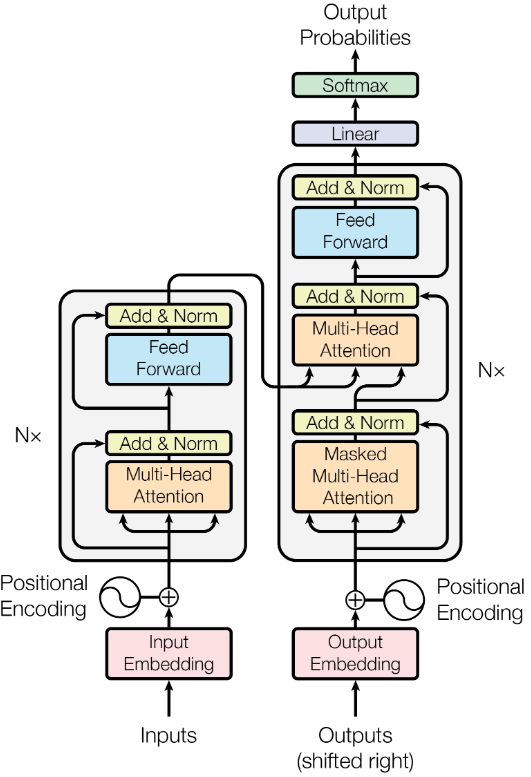}
\centering
\caption{The model architecture of a basic transformer as introduced in \cite{vaswani2017attention}}
\label{fig:bowDiagram}
\vskip 0mm
\end{figure}

The implementation of the summarisation mechanisms demonstrated in this work uses Sentence Transformer \cite{reimers2019sentencebert} (also known as sentence BERT or SBERT), a sentence vectorisation mechanism based on BERT \cite{devlin2018bert} (Sentence vectorisation would be covered separately in the next section). SBERT is chosen over other language models and encoders due to its high performance \cite{qiu2020pre} and convenience of implementation. Before explaining the sentence vectorisation mechanism executed by SBERT, it is worth mentioning the structure and brief working mechanism of BERT.

BERT, short for Bidirectional Encoder Representations from Transformers, is a renowned pre-trained language model based on the encoder part of the transformer architecture. A transformer includes both an encoder part and a decoder part, while BERT only uses the encoder part because its purpose is to generate a language model, not to perform a full-scale sequence conversion. As based on the encoder part of a transformer, BERT is bidirectional, so it reads the input sequence of words at once as a whole. This feature enables BERT to learn the "context" of each word through the surrounding words in a cost-effective way. It is trained by replacing 15\% of the word tokens in each sentence with a '[MASK]', a special token to train the model by trial-and-error for predicting the "correct" value for the '[MASK]'. The base BERT model is trained with 110 million parameters, and the large BERT model is trained with 340 million parameters. Applications built upon BERT show highly competent performances to various NLP tasks, including machine translation (it can still be utilised to deal with such tasks after some adjustments; see \cite{zhu2020incorporating} for details), question-answering \cite{rajpurkar2016squad}, and text classification \cite{sun2019fine}. 

\section{Sentence Vectorisation}

Sentence vectorisation in general could refer to a variety of text encoding mechanisms, from the bag-of-words representation to a full-scale semantic vectorisation such as SBERT \cite{reimers2019sentencebert}. The bag of words \cite{zhang2010understanding} model encodes a sentence (or a document) by counting how frequently each word in the pre-defined vocabulary appears in the sentence/document. It is widely used as a baseline model for various NLP tasks due to its ease of implementation, yet the vectors are usually very high in dimensionality because the size of the vector would be the same as that of the vocabulary, making it computationally expensive to run various calculations with the vectors. 

As mentioned, the Sentence Transformer (SBERT) framework to implement sentence vectorisation is used in this work. Sentence Transformer is a variant of BERT, putting its focus on finding and encoding the semantic meanings of sentences. The authors implemented the mechanism by adding a pooling layer at the end of the BERT transformer to generate a fixed-length vector that encodes the semantic textual similarity data from the BERT output. SBERT models are trained on 1 million sentence pairs created using text data from two sources: \cite{williams2017broad} and \cite{bowman2015large}. The database generated in \cite{williams2017broad} collects about 80,000 sentences for each of the ten genres, aiming to cover all possible genres in the English language. Among the ten, no genre directly overlaps with the domain of PP documents, but there is a genre called "government" believed to cover (at least partially) the legalese text of the documents. Also, database introduced in \cite{bowman2015large} contains image descriptions in pairs. The sentence pairs are either in an entailment relationship or in contradiction. Unlike the database in \cite{williams2017broad}, there are no specified genres to write the captions. The sentences represent generally used English sentences rather than PP documents written in legalese. The databases that the model is trained upon may not be task-specific to this work, yet the SBERT model itself shows strong results to encode English sentences in general \cite{reimers2019sentencebert}. Also considering the difficulty of collecting large enough task-specific data, using SBERT for sentence vectorisation for this work is believed to be justified. Sentences vectorised by Sentence Transformer can be similarity-tested using cosine similarity, Euclidean distance and Cityblock distance \cite{krause1973taxicab}.

\section{Principal Component Analysis}

For both clustering algorithms, Principal Component Analysis (PCA) \cite{pearson1901liii} is used for dimensionality reduction, as it seemed that Sentence Transformer's \cite{reimers2019sentencebert} 768-dimensional vector may affect the performance of the clustering algorithms. The details on how PCA was done for this work are explained in further detail in section 3.2.1. This section would briefly illustrate what PCA is and how it works. PCA aims to reduce the dimensionality of data if the high dimensionality harms the performance of analytical tools on the dataset. It removes irrelevant dimensions while keeping relevant dimensions in a few steps. First, it calculates eigenvalues and eigenvectors of the covariance matrix of the original data object. Then it sorts the eigenvalue - eigenvector pairs in descending order and select the first N eigenvectors, where N is a smaller number than the dimension of the original data. Finally, it transforms the original data object to be in N-dimension. There is no universally agreed rule-of-thumb to choose the best value for N, but a simple and typical way is to choose the number that achieves a pre-determined explained variance. Explained variance refers to the amount of information explained by the PCA model, measured by the ratio of accumulated variance over total variance. A typical value is 95\% \cite{qin1998determining}. Well-PCA-ed vectors are expected to be in a much lower dimension but include necessary data that the original dataset represented.

\section{Previous Approaches Explored}

This paper attempts to solve the problem of most internet users skipping Privacy Policy documents and mindlessly agreeing to them due to extensive reading requirements and ineffective website designs by summarising the documents based on machine-learning methods such as clustering. However, before implementing the aforementioned approach, I have explored multiple other methods to reach the same goal. The first is building a story generator using pre-trained language models such as BERT \cite{devlin2018bert} and GPT-2 \cite{radford2019language}, that warn users not reading certain PP clauses that are considered exceptionally important than others. These "exceptionally important" clauses were to be found on the internet using published articles and public studies regarding internet users' preferences for the privacy of personal data. Yet, the natural language generation algorithm was too complicated and targeted for addressing general NLP tasks such as answering questions using internet-based knowledge, not so relevant to the goal I targeted to achieve. It was demanding from the start to generate grammatically correct sentences, let alone generate sentences relevant to specific clauses worth noting to users. Also, a large task-specific corpus of data and a considerable amount of time to optimise applicable training algorithms were needed to implement the original idea, which seemed to be rather ambitious considering the limited resources in hand. 

After a number of extended experiments, I moved on to work on the same idea but with human-written template sentences, so that the system would not have to generate natural language "stories" from ground zero. To do this, classifying (or clustering) the PP sentences and mapping the classification results to \textit{n} risk levels was crucial, as the application was to be designed to pop out randomly chosen stories (template) to alarm users according to the risk level of each sentence. The assumed theory to this idea was that certain sections of PP documents such as third-party data sharing, data selling or marketing that are commonly recognised to be including sensitive data would be likely to cause more harm than other sections. However, not enough evidence from reliable sources was found to support the idea that the topics cause more harm to users. Most related articles such as \cite{cherivirala2016visualization} and \cite{liu2018towards} put more focus on accurately classifying the sentences into pre-determined classes, instead of identifying which classes would be more dangerous. 

\section{Related Works}
\cite{bergmann2007automatic} demonstrates a system that automatically clusters Privacy Policy preferences for different common use cases. For instance, users are commonly expected to \textit{make an order}, \textit{pay} and get their order \textit{shipped} to them at an online shopping website. For the website, the system clusters privacy preferences based on the three classes so that users do not have to read through the document and find clauses relevant to their actions. \cite{lu1978sentence} uses similarity histogram-based incremental clustering to cluster, order and find representative sentences on the internet to summarise them. The author uses ROUGE-1 \cite{lin2004rouge} to auto-evaluate the generated summaries.  \cite{gupta2012multi} works on multi-document summarisation by clustering a set of single-document summaries and selecting representative sentences to form a complete summary of multiple documents. \cite{moradi2018cibs} introduces a summarisation method for biomedical topics using clustering and text mining techniques, of which the method of clustering and extracting the most representative sentence is similar to that of previously mentioned works. \cite{yang2014enhancing} suggests that the traditional bag-of-words cosine similarity is no longer suitable for theme-based sentence clustering, proposing a ranking-based clustering framework. The authors demonstrate the effectiveness of their novel approach using the ROUGE summary evaluation method.

Summarisation techniques that do not use clustering algorithms as their major gear wheel of the mechanism are worth mentioning. Mainly, there have been introduced two ways of summarisation: extractive and abstractive. In extractive summarisation, the model considers the text summarisation problem as a sentence classification task to choose important ones out of many sentences in a document. The extractive summarisation idea is borrowed for this work due to its feasibility and effectiveness. Zhang et al. \cite{zhang2018neural} deal with summarisation as a task of deducing latent variables in a set of selected sentences. Zhou et al. \cite{zhou2018neural} suggest a score-based selective summarisation technique and show highly competent performances in the field of extractive summarisation. In abstractive summarisation, the model reads a document as a sequence of tokens and encodes it into a sequence of features, generating a summary token by token based on the feature sequence. Celikyilmaz et al. \cite{celikyilmaz2018deep} take a hierarchical attention mechanism to read input sequences using multiple encoders trained with reinforcement learning techniques. Narayan et al. \cite{narayan2018don} delve into extreme summarisation, shrinking the text into a single sentence using a Modified Convolutional Neural Network architecture.

Liu et al. \cite{liu2019text} improve the performance of extractive and abstractive summarisation methods by applying them to a pretrained BERT language model. The authors fine-tune the BERT model using each of the extractive and abstractive architectures and achieve relatively higher performance metrics than the previously mentioned leading summarisation algorithms including \cite{narayan2018don} and \cite{zhou2018neural}. Extractive summariser fine-tuned on the "large" BERT scored the best performance in terms of informativeness and fluency. 

\section{Background Chapter Summary}
This section explains how each section in this chapter directly relates to the aspects of this work. First, Reflections on my previous work \cite{choi2020worth} in section 2.1 became a motivation for this work. Next, the implementation of the summarisation mechanisms that this work explains are built upon the libraries that realise the algorithms explained earlier, including but not limited to: clustering (section 2.2.), transformers (section 2.3), sentence vectorisation using SBERT \cite{reimers2019sentencebert} (section 2.4) and principal component analysis (section 2.5). Clustering, SBERT sentence encoding (which is based on the encoder branch of the transformer architecture) and PCA are used to preprocess, encode and implement segregation of sentences in PP documents for summarisation. Section 2.6. shows the footprints of the exploration to find the most efficient method to solve the ultimate goal of this work to facilitate the readings of PP documents for ordinary users. Finally, related works elaborated in section 2.7 inspired me to get relevant ideas and understand the domain of text summarisation.

\chapter{Methodology}
   
\section{Data collection}

This paper compares two different summarisation techniques to find a better solution to the Privacy Policy summarisation task: K-means clustering and PDC (pre-determined centroid) clustering. Therefore, a different dataset is used for each of the two clustering models. As opposed to my previous work \cite{choi2020worth} where multiple personas were used for a personalised experience in reading Privacy Policy documents easily and fast, the PP clause summarisation mechanism proposed in this paper uses the same dataset but does not succeed in the idea of using personas for personalisation. Instead, a user can choose the length of the summary they would like to get. Based on the length parameter of the system, it covers from sensitive and careful users to uninterested and resistant users. Therefore, the 2382 sentence chunks used in the work that are generally composed of 2 to 4 sentences, are split by full stops to become 5778 sentences - each of which is composed of 15 words on average. Due to the nature of the sentences that they are manually collected to be read by human workers for data labelling, the sentences were somewhat pre-processed to be in a human-friendly state. For instance, the company names in the sentences are replaced with first-person perspective words such as "we", "us" and "ours". Headings, short bullet points and links are removed as well, for the likewise reason. Table \ref{tab:olddata} briefly shows a snippet of the data used in my previous work \cite{choi2020worth}.

\begin{table}[]
\begin{tabular}{|l|l|l|}
\hline
\textbf{company} & \multicolumn{1}{c|}{\textbf{clause}}                                                                                                                                                                                                                                                                              & \multicolumn{1}{c|}{\textbf{class}} \\ \hline
facebook         & \begin{tabular}[c]{@{}l@{}}Information and content you provide. We collect the content, \\ communications and other information you provide when you\\ use our Products, including when you sign up for an account, \\ ... Learn more about how you can control who can see the \\ things you share.\end{tabular} & standard\_trivial                   \\ \hline
facebook         & \begin{tabular}[c]{@{}l@{}}We collect information about how you use our Products, \\ such as the types of content that you view or engage with, \\ the features you use, the actions you take, the people or ... \\ We also collect information about how you use features \\ such as our camera.\end{tabular}    & worth\_reading                      \\ \hline
facebook         & \begin{tabular}[c]{@{}l@{}}If you have it turned on, we use face recognition technology \\ to recognise you in photos, videos and camera experiences. \\ The face recognition templates that we create are data with\\  special protections under EU law. Learn more about ...\end{tabular}                       & worth\_reading                      \\ \hline
facebook         & \begin{tabular}[c]{@{}l@{}}We use your information to enhance the safety and security\\ of our products and services.\end{tabular}                                                                                                                                                                                & standard\_trivial                   \\ \hline
\end{tabular}
\caption{A snippet of the collected data used in previous work \cite{choi2020worth}}
\label{tab:olddata}
\end{table}

For the PDC clustering model, EU's \cite{wodford_2019} instruction on writing a GDPR-compliant privacy policy (or notice) is used as a source to extract relevant data to train the model. The article specifies 14 essential components that a compliant  PP must include, such as the types and contents of the data the company collects, how to store them, for what purposes would the company use the data and so on. Table \ref{tab:14topics} shows all the 14 topics instructed in \cite{wodford_2019}. The article includes multiple sample clauses that belong to each of the 14 topics that the EU believes to be GDPR-compliant. For each topic, the sentences are combined to form a single sentence. The combination is manually done and the purpose of it is to make the representative sentence for each topic include as many meaningful words as possible. The sentences produced by the combining behaviour are not expected to be grammatically complete and clear to human readers, but by doing so all the potential key information about the topic could be included in each "gold standard" sentence. It is empirically realised that the sentence combination makes it easier for sentence vectors to spot the closest centroid, not only to the machine but also to a human eye. An example of the combined sentences is \textit{"we send you information about products and services you might like recommend marketing third party use opt out later right to stop no longer wish marketing purposes"}. For clustering in the later steps, the 14 combined sentences are encoded to 14 vectors and saved into a local disk, using Sentence Transformers \cite{reimers2019sentencebert}. 

\begin{table}[!h]
\centering
\begin{tabular}{|c|c|}
	\hline
	& Topic \\
	\hline\hline
	1 & What data do we collect? \\
	\hline
	2 & How do we collect your data?    \\
	\hline
	3 & How will we use your data?     \\
	\hline
	4 & How do we store your data?      \\
	\hline
	5 & Marketing  \\
	\hline
	6 & What are your data protection rights?    \\
	\hline
	7 & What are cookies?       \\
	\hline
	8 & How do we use cookies?   \\
	\hline
	9 & What types of cookies do we use?  \\
	\hline
	10 & How to manage your cookies  \\
	\hline
	11 & Privacy policies of other websites     \\
	\hline
	12 & Changes to our privacy policy   \\
	\hline
	13 & How to contact us    \\
	\hline
	14 & How to contact the appropriate authorities \\
	\hline
\end{tabular}
\caption{The 14 essential topics specified by GDPR that must be included in Privace Notice of companies located in Europe.}
\label{tab:14topics}
\end{table}

\section{Clustering}

For easier understanding and shorter referring representations, the PP sentences clustering mechanism based on K-means clustering will be denoted as \textit{SumA}, and the clustering model based on PDC (pre-determined centroids) clustering will be denoted as \textit{SumB}. 

Unlike \textit{SumB} whose centroids (therefore representative sentences for 14 essential topics) are known, centroids for \textit{SumA} is not known and finding them is the primary goal of performing the clustering algorithm. The clustering for \textit{SumA} is implemented as follows: First, the sentences are encoded to vectors using the same BERT-based sentence-vectorisation library \cite{reimers2019sentencebert}. Then, the vectors are clustered via Scikit-Learn's \cite{scikitLearn} implementation of K-means clustering algorithm \cite{macqueen1967some} where the value for \textit{k} (number of clusters) is 14, given that there are 14 essential topics specified in the GDPR article \cite{gdpr_2019}. 

\subsection{Principal Component Analysis}

Before running the K-means algorithm on the training dataset, the dimensionality of the sentence vectors was reduced by Principal Component Analysis (PCA). Each of the original sentence vectors had 768 features, which made it rather difficult to assign the vectors to 14 clusters due to the high dimensionality. The optimal value for the number of principal components was chosen by two methods: 1) Silhouette score maximisation and 2) explained variance ratio. \\

\begin{figure}[!tbp]
  \centering
  \begin{subfigure}[b]{0.49\textwidth}
    \includegraphics[width=\textwidth]{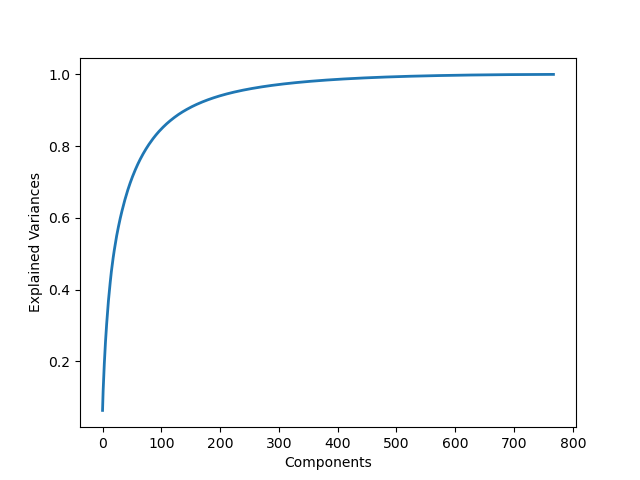}
    \caption{Explained variance \textit{n\_comp}.}
    \label{fig:pca_variance}
  \end{subfigure}
  \hfill
  \begin{subfigure}[b]{0.49\textwidth}
    \includegraphics[width=\textwidth]{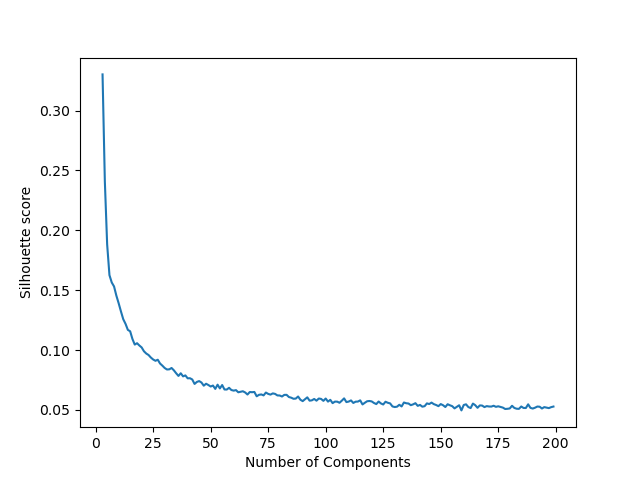}
    \caption{Silhouette score by \textit{n\_comp}.}
    \label{fig:pca_sil}
  \end{subfigure}
  \caption{Explained variance (left) and Silhouette score (right) by the number of principal components.}
\end{figure}

Explained variance ratio is a metric to calculate the variance attributed by principal components, a way to decide the optimal number of principal components. \cite{hong2020selecting} suggests that a typical percentage value for explained variance to choose the number of principal components would be around 90\%. When the number of components was 140, 90\% of the variance was explained, and 85\% was explained with 100 principal components. Hence, for this criteria 140 and 100 are chosen as the candidates for optimal \textit{n\_comp} value. Figure \ref{fig:pca_variance} illustrates how the explained variances increase as the number of principal components grows. \\

Another approach used to determine the optimal value for \textit{n\_comp} is plotting the Silhouette score recorded by different numbers of components. Silhouette score \cite{rousseeuw1987Silhouettes} is a metric to evaluate the clustering result by considering the mean of intra-cluster distances and mean of nearest-cluster distances. The score is a value between -1 and 1, where 1 means that members are well assigned to relevant clusters and there are no major overlaps. A value near 0 represents that the clusters are very close to other clusters with large overlapping areas. Figure \ref{fig:pca_sil} shows how the Silhouette score drops drastically as the number of components increases. The Silhouette score was maximised when there were only 3 principal components, and it never bounced back. To test out the clustering algorithm with two different values with relatively higher Silhouette scores, 3 and 10 are chosen as candidates for deciding the optimal \textit{n\_comp} value.

\subsection{Clustering algorithms explored}

10 clustering algorithms are investigated and used to find out the most effective model for PP document sentence clustering. The names of the tried algorithms are: \textit{Kmeans}, \textit{mini batch kmeans}, \textit{DBSCAN}, \textit{HDBSCAN}, \textit{OPTICS}, \textit{Agglomerative}, \textit{Gaussian}, \textit{BIRCH}, \textit{Affinity} and \textit{Meanshift}. The mini-batch K-means algorithm is proposed by Sculley \cite{sculley2010web}, where the training dataset is sliced into smaller batches and fed to the training model. Usually, the algorithm is used for a much larger dataset, though it is used in this work to test out the algorithm on this model for an experiment. DBSCAN \cite{hahsler2019dbscan} is short for Density-Based Spatial Clustering of Applications with Noise. It clusters data points based on the density of areas where data points are placed on. It is robust to outliers due to its focus on spatial densities for clustering, and it also does not require the user to input the number of clusters. HDBSCAN \cite{mcinnes2017hdbscan} is a hierarchical variant of DBSCAN which puts more aim on areas with a higher density. 

The OPTICS algorithm \cite{ankerst1999optics} is an abbreviation of Ordering Points To Identify Cluster Structure. It extends the DBSCAN algorithm \cite{hahsler2019dbscan} by taking two more key distance concepts into account: core distance and reachability distance. Core distance refers to a distance value required for a data point to be selected as a centroid candidate. Reachability distance is calculated between two data points p and q, by performing $max(x,y)$ where $x$ is the core distance of p and $y$ is the Euclidean distance between p and q, and it is the mainly used distance value for OPTICS clustering. OPTICS is distanced from other clustering algorithms in that its functional nature is in displaying a Reachability distance plot for the user to choose to do further clustering and not in explicitly clustering the given data points. \\

Agglomerative clustering \cite{mullner2011modern} is a well-known hierarchical clustering algorithm. It starts by regarding all data points as singleton clusters and merging them to form larger clusters by their similarities until there is no larger cluster left to be merged. Classic metrics to measure similarities are Euclidean distance and Manhattan \cite{krause1973taxicab} distance. Agglomerative hierarchical clustering does not require the user to input the number of clusters. Also, it is capable of producing an ordering of data points which may be used to provide a better understanding of the given dataset. However, it is unable to undo clustering behaviours of early stages to get a better result as other clustering algorithms do. 

Gaussian mixture model clustering \cite{yang2012robust} aims to address K-means clustering's issue of force-fitting data points to circular clusters. Compared to K-means' circular distance and DBSCAN's spatial density, GMM clustering utilises multi-dimensional Gaussian probability distributions to assign data points to the most likely clusters. It can find and register data points to non-circular clusters. BIRCH \cite{zhang1996birch} is short for Balanced Iterative Reducing and Clustering Hierarchies, a hierarchical clustering algorithm often used with large datasets. Unlike K-means or Agglomerative clustering, BIRCH does not cluster the entire data points at once. Instead, it converts the dataset into a tree-like data structure (called a CF tree), starting by summarising smaller subsets and escalating the hierarchy to finally cluster the total data points. BIRCH only requires a single scan of the dataset to start clustering therefore it is executed fast, as well as is a scalable clustering solution due to its hierarchical structure.

Affinity Propagation clustering \cite{dueck2009affinity} is a clustering algorithm that decides the optimal number of clusters automatically based on two metrics: similarity and preference. In Affinity Propagation, all data points are considered as nodes in a network, where all nodes can be regarded as potential centroids. It takes the similarities matrix of the data nodes which represents how well an exemplar candidate data point $x_k$ can represent a "member" data point $x_i$ compared to other centroid candidate data points. The algorithm also requires preferences, which tell how "appropriate" it would be to choose a data point $x_k$ as a member data point $x_i$'s exemplar, compared to other member data points. 

The Mean shift \cite{derpanis2005mean} algorithm clusters data points by shifting windows of pre-defined radius (or "bandwidth") to the mean of the data points that it is covering. The window keeps moving to a region of a higher density until it fails to find a denser area and converges. The window generation and shifting are done for each data point to find the best clustering result. Mean Shift clustering is convenient in that it does not require us to specify the value for the number of clusters, and that it only takes a single parameter: bandwidth. Yet, due to its structural limitation of having to perform mean-shifting actions on a large number of windows (as many as the number of data points), it is computationally expensive. Also, choosing the optimal value for bandwidth could often be rather difficult \cite{wang2004image}.  

\subsection{Silhouette scores by clustering algorithm}

\begin{figure}[!tbp]
  \centering
  \subfloat[Kmeans clustering result visualised]{\includegraphics[width=0.5\textwidth]{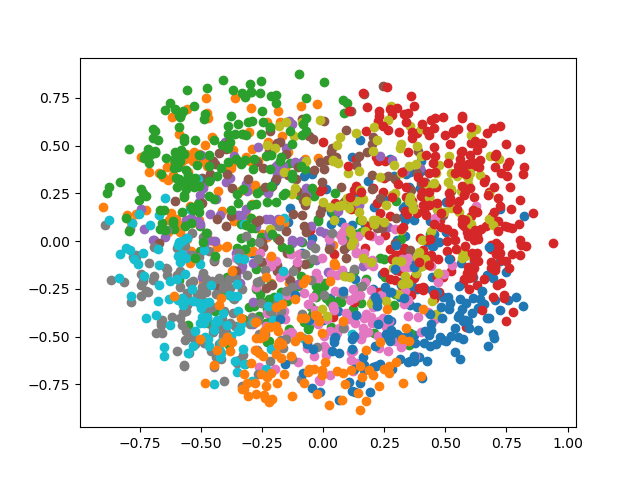}\label{fig:kmeans_plot}}
  \hfill
  \subfloat[Affinity clustering result visualised.]{\includegraphics[width=0.5\textwidth]{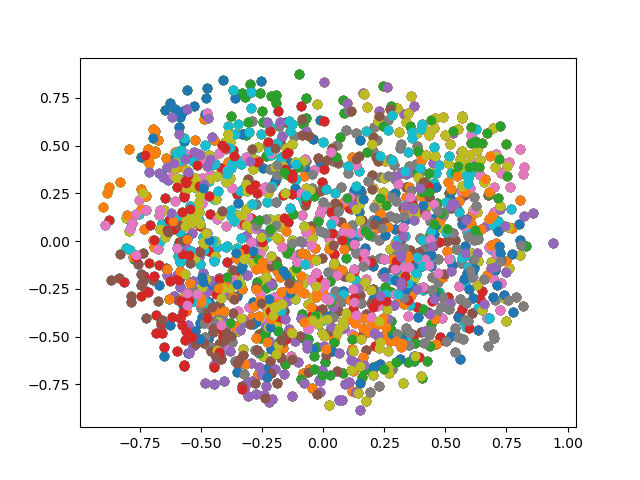}\label{fig:affinity_plot}}
  \caption{Comparison of clustering results made by Kmeans and Affinity.}
\end{figure}

Table \ref{tab:performances} compares the Silhouette scores of clustering results generated by the 10 clustering algorithms, by 4 different \textit{n\_comp} values 3, 10, 100 and 140. As expected, smaller \textit{n\_comp} values resulted in higher Silhouette scores regardless of the choice of algorithm (excluding the algorithms that recorded negative Silhouette scores). Affinity recorded the highest Silhouette score of 0.3455 with 3 principal components. However, Affinity is an algorithm that does not take the number of clusters as a parameter. It rather decides the optimal number of clusters itself, and the value it used here is 42 - meaning that the training dataset was clustered into 42 clusters. When k=42, the Silhouette score brought by Kmeans is 0.3535, beating Affinity's 0.3455 by a margin of 0.008. The difference in the Silhouette score between the two results is quite trivial (0.008 on a scale of -1 to 1), therefore it may be difficult to claim that Kmeans is superior to Affinity for the clustering task. However, such a statement is only valid if trying many different values for cluster numbers is allowed to reach the maximum Silhouette score. The number of clusters in this work is fixed to 14 for feasibility because specifying the value otherwise or using a larger number of clusters to maximise the Silhouette score would require the clusters to be mapped into 14 afterwards. This job would demand a considerable amount of manual work which would be counterproductive considering the time budget allocated for this work. It should be considered more reasonable to use K-means because it achieves a fairly solid Silhouette score of 0.3301 (the difference between Affinity is 0.0154 which is still insignificant) even when the number of clusters is specified as 14. This is the reason that the Silhouette scores of K-means are highlighted in bold even though the scores of Affinity are apparently higher. Figure \ref{fig:kmeans_plot} and \ref{fig:affinity_plot} visualise the clustering results performed by K-means and Affinity clustering algorithms, given that k=14 and \textit{n\_comp}=10. The parameter k=14 is not applied to the Affinity algorithm as explained. The clusters in Figure \ref{fig:kmeans_plot} are easier to distinguish compared to the clusters in Figure \ref{fig:affinity_plot}. 

Supported by the Silhouette scores and reasons stated above, the K-means clustering algorithm was chosen to be used as a representative algorithm of the conventional clustering algorithms, to be compared against PDC clustering in the later \textit{Evaluation} chapter. After clustering the sentence vectors using K-means, a sentence vector closest to the centroid (the mean vector of all data points in a cluster) is noted as a "pseudo-centroid" and saved into a disk space for later use.

\vskip 5mm
\begin{table}[h!]
\centering
\begin{tabular}{|c|r|r|r|r|}
\hline
Algorithm & \multicolumn{1}{c|}{\shortstack{Silhouette \\ (n\_comp=3)}} & \multicolumn{1}{c|}{\shortstack{Silhouette \\ (n\_comp=10)}} & \multicolumn{1}{c|}{\shortstack{Silhouette \\ (n\_comp=100)}} & \multicolumn{1}{c|}{\shortstack{Silhouette \\ (n\_comp=140)}}\\ \hline\hline
Kmeans              & \textbf{0.3301} & \textbf{0.1342} & \textbf{0.0562} & \textbf{0.0528}  \\ \hline
Minibatch\_kmeans   & 0.2977    & 0.1187    & 0.0518    & 0.0513        \\ \hline
DBScan              & FAILED    & FAILED    & FAILED    & FAILED        \\ \hline
HDBScan             & -0.1816   & -0.1641   & -0.0522   & -0.0512       \\ \hline
Optics              & -0.3481   & FAILED    & 0.0066    & 0.0066        \\ \hline
Agglomerative       & 0.2755    & 0.0798    & 0.035     & 0.0272        \\ \hline
Gaussian            & 0.3170    & 0.1071    & 0.0542    & 0.0519        \\ \hline
Birch               & 0.2747    & 0.0798    & 0.0351    & 0.0272        \\ \hline
Affinity            & 0.3455    & 0.1121    & 0.0596    & 0.0567        \\ \hline
Meanshift           & 0.3049    & FAILED    & FAILED    & FAILED        \\ \hline
\end{tabular}
\caption{Silhouette scores of different clustering algorithms after training. Sentence vectors were reduced in dimensionality using PCA with four different \textit{n\_comp} values: 3, 10, 100 and 140. The indicator "FAILED" appears on the clustering results for algorithms that do not require the parameter \textit{number of clusters} (DBSCAN, HDBSCAN, Optics, Meahshift and Affinity) when the algorithm fails to segregate the data points into 2 or more clusters and simply returns the entire dataset as a large single cluster.}
\label{tab:performances}
\end{table}

\begin{figure}[!tbp]
  \centering
  \subfloat[PDC clustering result with 3 PC]{\includegraphics[width=0.5\textwidth]{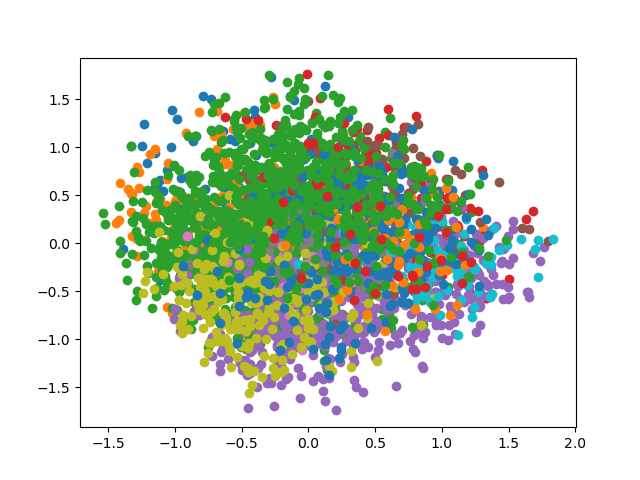}\label{fig:pdc_np3}}
  \hfill
  \subfloat[PDC clustering result with 100 PC]{\includegraphics[width=0.5\textwidth]{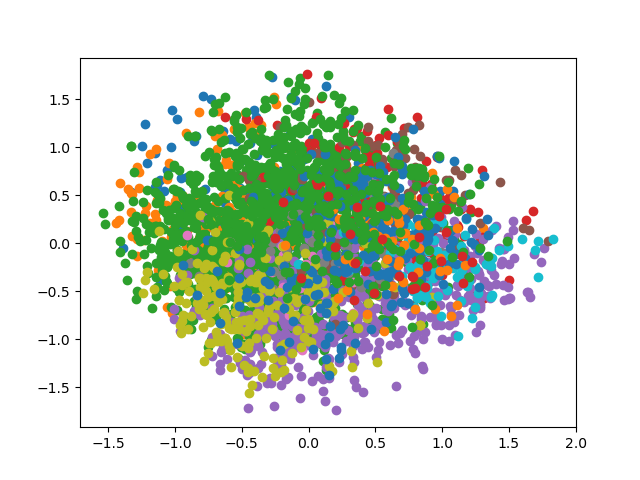}\label{fig:pdc_np140}}
  \caption{Comparison of clustering results generated by the PDC-clustering algorithm with two separate \textit{n\_comp} values 3 and 100.}
\end{figure}

For \textit{SumB}, the 14 vectorised sentences acquired in the previous section are used as pre-determined centroids of the clustering algorithm. The major difference between PDC clustering and K-means is that in PDC-clustering, the centroids are already defined so there is no need to fit the data to the training dataset to find cluster centres. Instead, the PDC module calculates the Euclidean distance between the pre-defined centroids and data points and assigns them to the closest cluster centres. The PDC-clustering algorithm got a Silhouette score close to zero (0.0007). A Silhouette score close to zero implies that a big proportion of clusters overlap each other, and it was the case here. It was not so easy to distinguish the clusters from a human eye. Figures \ref{fig:pdc_np3} and \ref{fig:pdc_np140} display the visualised clustering result produced by the algorithm. The reason that the general Silhouette score of the PDC-clustering model was much worse than that of the K-means model is that the cluster centres are pre-determined and cannot be modified. K-means updates the position of centroids in a way that minimises the sum of distance from the centroid to its member data points. Therefore, it is designed to prioritise reducing the sum over anything else. The average distance between inter-cluster data points and that between intra-cluster data points determine the Silhouette score, so K-means clustering is better optimised in getting a higher Silhouette score than PDC. In contrast, the primary aim of PDC-clustering in this work is to find a number of data points near pre-defined cluster centres and analyse the distribution. It is not optimised to reduce the sum of distance, simply because it is not an objective of the algorithm. How the PDC clustering algorithm performs for the ultimate task of finding PP sentences relevant to the 14 topics would be covered in the next chapter. Therefore, as can be seen in both figures \ref{fig:pdc_np3} and \ref{fig:pdc_np140}, PDC-clustering was unsuccessful in segregating the data points clear to human eyes. Also, the two figures seemed almost identical even though the number of principal components was quite different (3 and 100), therefore it is found that tweaking the number of principal components did not produce any meaningful change to the clustering result of K-means.

\section{Summariser Implementation}

\begin{figure}[h!]
\includegraphics[width=\textwidth]{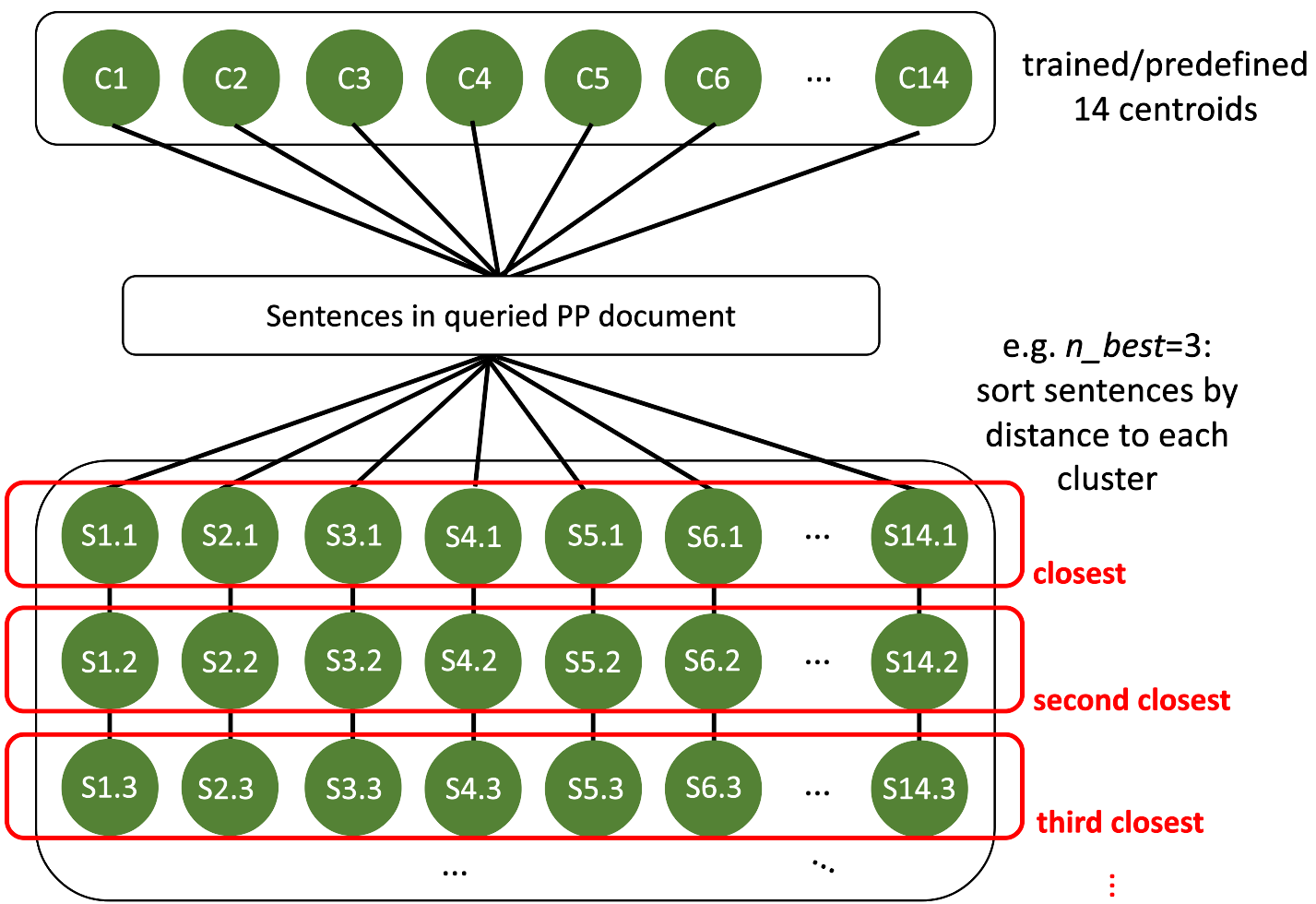}
\centering
    \caption{A brief visual representation of how the search is processed.}
\label{fig:summariser}
\vskip 5mm
\end{figure}

The summariser application is designed to take two inputs: 1) the URL where the Privacy Policy document is located, and 2) \textit{n\_best} which represents the number of sentences the user would like to get, sorted by distance to each centroid. This applies to both summarisers implemented upon PDC clustering and the K-means clustering algorithm. As briefly mentioned in section 3.1, users can choose the length of the PP document summary based on their sensitivity level to data privacy. For instance, the most sensitive persona mentioned in \cite{choi2020worth} is Alice, who is naturally more motivated to read more about how her data would be used, stored and so on. A user like Alice would choose higher \textit{n\_best} values, 10 for instance. In contrast, users like Charlie (as in \cite{choi2020worth}, the persona with the lowest data privacy sensitivity with lots of prior knowledge) would choose much lower numbers like 1 or 2 to keep their summaries as short as possible. Diagram \ref{fig:summariser} illustrates how the summariser processes the query. When it is given the URL, the web-scraping module scrapes all the text data stored in the HTML body, removes unsupported punctuation marks, splits the text by full stops and returns the pre-processed sentences into a list. As soon as the list returns, it calculates the Euclidean distance between each centroid to the list of sentences and sorts the distances. Here, if the summarisation mode is set to \textit{kmeans}, then the centroids generated by training the k-means model would be used for the distance calculation. If the mode is \textit{pdc}, then the GDPR-referred pre-defined centroids would be used to calculate the distances.

\begin{sidewaysfigure}
    \centering
    \includegraphics[width=1.1\textwidth]{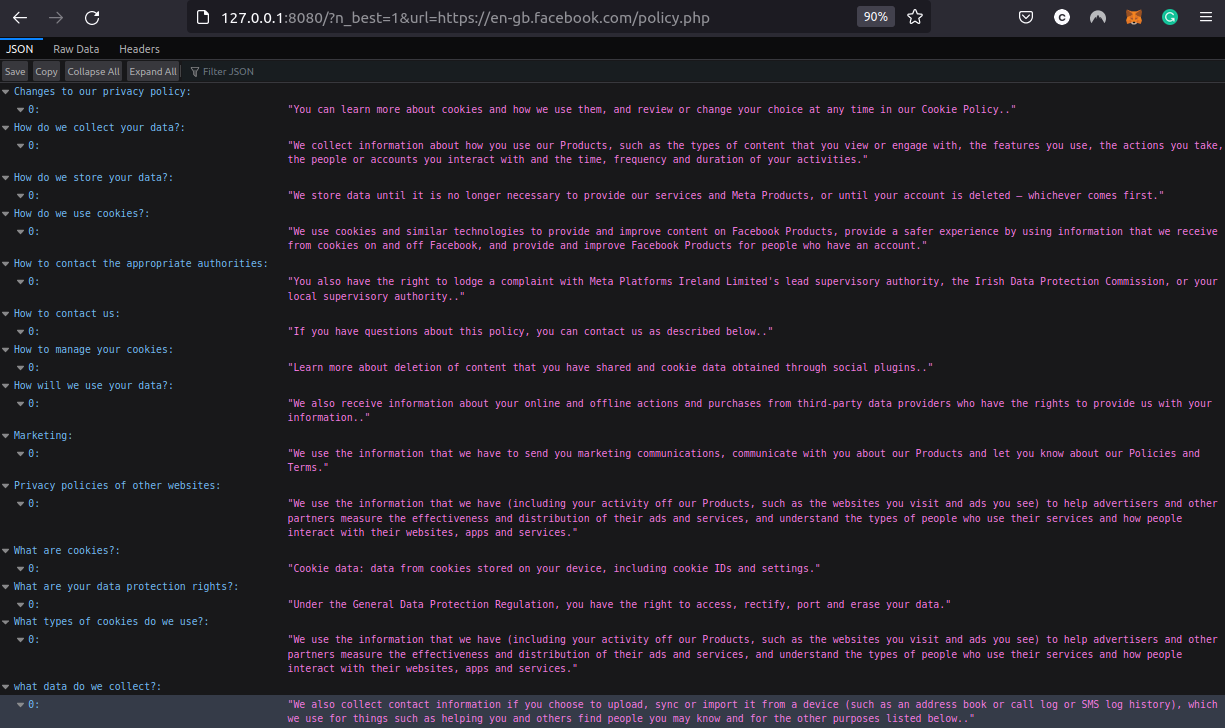}
    \caption{An example summary of Meta (former Facebook)'s Privacy Policy using the PDC-clustering based summariser where \textit{n\_best}=1}
    \label{fig:facebooksummary}
\end{sidewaysfigure}

When the distance calculation step is finished, the summariser sorts the sentences by distance to each centroid, formatting them to be in a human-readable JSON. Finally, given the \textit{n\_best} value, it decides which sentences should be discarded and which should not. For instance, if \textit{n\_best} is 1, all sentences except for those who are the closest to each topic would be discarded, and only the 14 sentences will be returned. Figure \ref{fig:facebooksummary} shows the returning JSON object acquired by running the summariser on Meta's (former Facebook) Privacy Policy document. Here, \textit{n\_best} is set to 1 for simplicity. The number of clauses given after pre-processing is 260 which is reduced to 14 by the summarisation mechanism. This is a 94.6\% reduction, expected to dramatically reduce users' time reading the policy and understanding its essential contents of it. Yet, the rate of reduction in the number of sentences is easily affected by the parameter \textit{n\_best}, and also the summariser would be of no use if it merely chooses random sentences from the original document. Therefore, maintaining a good quality of the summary as well as adequately reducing the length would be the key aspect of the summariser to insist that the software is useful. A detailed evaluation of the quality of the summary will continue in the next chapter. 

\chapter{Evaluation}

\section{Methodology}

This section explains two evaluation methods used to demonstrate the usefulness of the PDC-clustered PP document summariser, where the evaluation metrics scored by the K-means based summariser would be used as a benchmark. The first method is Sum of Squared Distance (SSD), and the second is the ROUGE \cite{lin2004rouge} summariser evaluation metric. SSD is chosen over average Euclidean distance, and this is because the difference in Euclidean distances between two "somewhat relevant" sentence vectors (all sentences are somewhat related to each other in that they semantically represent privacy notices, regardless of their topic) was rather small, requiring a tool to magnify the distances for easier recognition.

\subsection{Sum of Squared Distance (SSD)}

\begin{figure}[h!]
\centering
\includegraphics[width=0.9\textwidth]{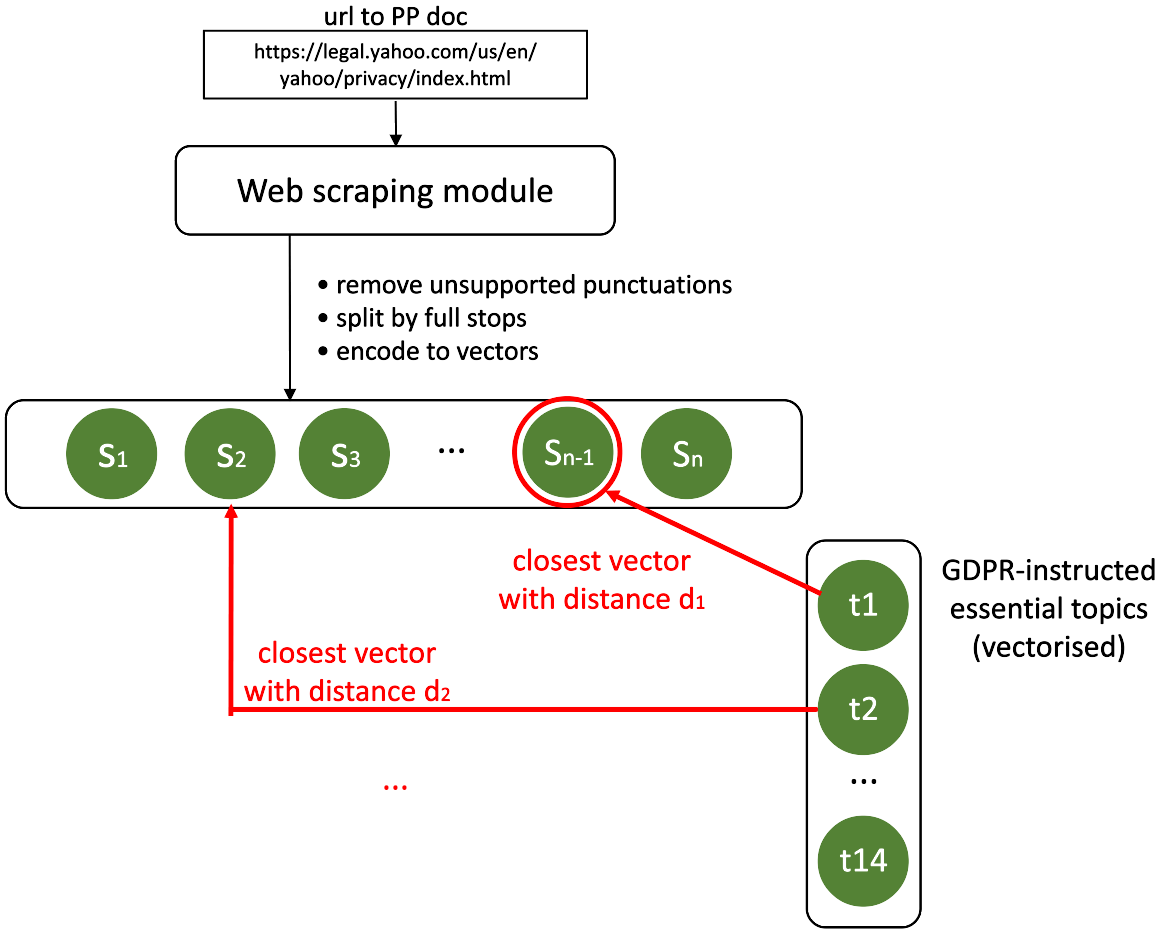}
    \caption{A workflow diagram on how the SSD evaluation metric is calculated.}
\label{fig:ssd_eval}
\end{figure}

The 14 topics specified in GDPR's instruction \cite{wodford_2019} illustrated in table \ref{tab:14topics} are used as the pivot to calculate distances. First, the topics (topic headers in table \ref{tab:14topics}) are encoded to vectors with the pre-trained Sentence Transformer model. Then for each topic sentence vector, the squared Euclidean distance between each sentence vector in the generated summary is calculated and the smallest distance value will be assigned to the corresponding topic. The assigned squared Euclidean distances will be summed up to a single number, and this sum would be returned as the evaluation score. To understand the meaning of the numbers, the evaluation metric is used on 4 summaries for comparison: 1) randomly generated 14 sentences \cite{randomwordgenerator2022} 2) summary generated by the summarisation mechanism based on PDC-clustering, 3) summary generated by K-means clustering based summariser and 4) the 14 combined sentences used to train the PDC clustering algorithm. It is expected that 1) will show the worst result \textit{i.e.} largest SSD (as this model is a random baseline to contrast with the other summariser models), and 4) would record the most competent result (smallest SSD). Before running the evaluation it was not sure which of the two summarisers would perform better \textit{i.e.} return lower SSD value. Privacy Policy documents from the 50 most-visited websites in the world (disregarding academically inappropriate websites) are used to average out the numbers so that the evaluation could be fair enough to all summariser models. Figure \ref{fig:ssd_eval} illustrates the flow of SSD evaluation on the generated summaries.

\subsection{ROUGE}
For ROUGE, human-annotated summaries were required as the gold standard to compare against the generated summary. For a fair comparison along with the SSD evaluation, the GDPR combined sample sentences (as explained in section 3.1.) are used as the reference for the evaluation of the generated sentences. The same list of 50 URLs introduced in the previous section for SSD evaluation is used to find PP documents. For each reference sentence in each PP document, ROUGE-N with \textit{max\_n}=2, ROUGE-L and ROUGE-W are calculated and visualised for further analysis. ROUGE-N finds matching N-grams between the hypothesis sentence and the reference summary for different N values. ROUGE-L looks for the longest matching subsequence between the two summary sentences. ROUGE-W considers the weighted longest common subsequences between the two sentences. Here, the value for W is to be determined and the default value is 1.0. For ROUGE evaluation, I used the py-rouge package \cite{pyrouge} as it claims to present the same results as the official Perl implementation of ROUGE. The Rouge score is calculated by each topic, referring to the relevant combined sample sentence.

\section{Results}

\subsection{Sum of Squared Distance (SSD)}
To help understand the meaning of squared distance between two vectors, Table \ref{tab:sqdistComp} displays a comparison of summary sentences selected by each of the four summarisation models (PDC, Kmeans, GDPR and Random baseline). To be exact, "GDPR" on the table does not refer to a summarisation model but a document of the sample combined sentences that represent the key meanings of the 14 essential GDPR topics. Likewise, "Random baseline" is not a model but a document including 14 randomly generated sentences. Yet, for a simpler explanation, the documents (GDPR and Random baseline) would be addressed as "models" instead of documents. On the table (table \ref{tab:sqdistComp}), the summary sentence found by the random baseline model is clearly distanced from the other three, as the sentence does not share any meaning with the topic keyword "marketing". Therefore, the squared distance of the sentence to the topic vector is found to be 21.5765. The summary sentence indicated by the GDPR model has the closest squared distance of 13.6310, yet the distance difference between GDPR and the other two (PDC and Kmeans) is not as large as that between GDPR and Random baseline. The summary sentences found by the PDC model and the Kmeans model fairly talk about marketing-related content, but the sentence selected by the PDC model is found to be closer to the GDPR model's selected sentence. This indicates that the PDC model found a summary sentence that has closer meaning to the GDPR standard sentence for this topic. Accordingly, the difference of 0.627 between the two squared distance values generated by each of the PDC and K-means models is meaningful enough to judge that which chosen sentence is a "better" summary. As elaborated in the previous section, SSD is calculated per summarisation model by adding up the squared distance values (as in the rightmost column in table \ref{tab:sqdistComp}) computed by squaring the distance between the nearest sentence vector and the relevant topic vector.

\begin{figure}[h!]
\includegraphics[width=\textwidth]{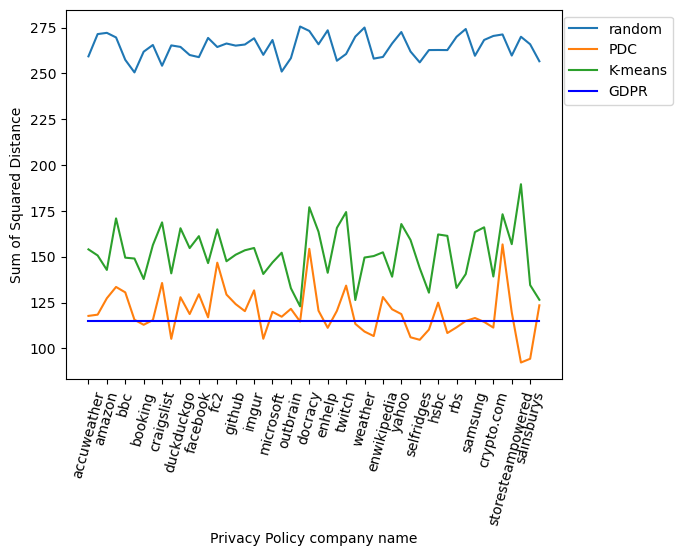}
\centering
    \caption{A visual representation of the SSD evaluation results. Only half of the company names are displayed on the X axis for better readability.}
\label{fig:ssd_plot}
\vskip 3mm
\end{figure}

Table \ref{tab:ssdresults} shows the SSD evaluation scores of the four summaries for each company's PP document, including the mean and the standard deviation. GDPR-instructed (SSD from GDPR topics to GDPR sample PP document is fixed as there is no change to the document) sample sentences showed the smallest SSD due to the design of the evaluation. As expected, the mean SSD of randomly generated sentences was more than twice as large as that of GDPR sample sentences (the specific value was 264.52), meaning that the semantic meanings of the random sentences were the least similar to the gold standard topics. The mean SSD of summary sentences generated by the K-means based summariser was 152.06, beaten by the PDC-clustering based summariser by a margin of 27\%. The standard deviation of the SSD scores recorded by the two summarisers was 14.28 and 12.51. This means that the SSD of PDC's worst summary (132.19) was still smaller \textit{i.e.} better than that of the best summary created by the K-means model (137.78). This is well represented in Figure \ref{fig:ssd_plot}. The SSD values on the green line (K-means) never reach the values on the orange line (PDC). Therefore, it is reasonable to say that in terms of SSD evaluation, the PDC-clustering based summariser outperforms the K-means based one. On some occasions, summary sentences generated by the PDC-clustering based model indicated lower SSD values than that of GDPR's. This is believed to be attributed to the fact that the GDPR "sample sentences" used in this experiment are not the exact sentences suggested in GDPR Privacy Notice instruction, but are combinations of GDPR sentences as explained in section 3.1. As the combined sentences are much longer than unedited sample sentences, it is likely that the distance between the vectorised topic and the sentence vector could have been artificially magnified.

\begin{table}[]
\begin{tabular}{|lll|}
\hline
\multicolumn{3}{|c|}{\textbf{GDPR Topic: Marketing}}                                                                                                                                                                                                                                                                                                                                                                                 \\ \hline
\multicolumn{1}{|c|}{\textbf{Model}}                                                     & \multicolumn{1}{c|}{\textbf{Sentence found}}                                                                                                                                                                                                                                            & \multicolumn{1}{c|}{\textbf{Sq. distance}} \\ \hline
\multicolumn{1}{|l|}{\textbf{PDC}}                                                       & \multicolumn{1}{l|}{\begin{tabular}[c]{@{}l@{}}Information on how you can opt out of selling your \\ information with these vendors for purposes beyond\\ the advertising services they provide us is provided \\ below (“Opt out of sale of your Personal Information”).\end{tabular}} & 14.5590                                    \\ \hline
\multicolumn{1}{|l|}{\textbf{Kmeans}}                                                    & \multicolumn{1}{l|}{\begin{tabular}[c]{@{}l@{}}We do not control these third-party entities, products, \\ and services, and they may collect, use, process, transmit,\\ and disclose your information.\end{tabular}}                                                                    & 15.1860                                    \\ \hline
\multicolumn{1}{|l|}{\textbf{GDPR}}                                                      & \multicolumn{1}{l|}{\begin{tabular}[c]{@{}l@{}}we send you information about products and services you\\ might like recommend marketing third party use opt out \\ later right to stop no longer wish marketing purposes.\end{tabular}}                                                 & 13.6310                                    \\ \hline
\multicolumn{1}{|l|}{\textbf{\begin{tabular}[c]{@{}l@{}}Random\\ baseline\end{tabular}}} & \multicolumn{1}{l|}{\begin{tabular}[c]{@{}l@{}}The toddler’s endless tantrum caused the entire \\ plane anxiety.\end{tabular}}                                                                                                                                                          & 21.5765                                    \\ \hline
\end{tabular}
\caption{A comparison of summary sentences selected by each summarisation model and the squared distance of each sentence to the topic vector "Marketing". The URL of the PP document used for the analysis is as follows: https://weather.com/en-US/twc/privacy-policy}
\label{tab:sqdistComp}
\end{table}

As mentioned in section 3.2, the Silhouette score made by the K-means clustering based summariser model (0.3301) is much larger than that performed by the model based on PDC-clustering (0.0007), representing that the data points are well-assigned to appropriate clusters. However, even though it may be true that the data points are segregated better with K-means, the evaluation performed in this chapter is aimed at finding sentences in a PP document that contains essential information required by GDPR. The centroids calculated by the K-means model may be somewhat separated from the 14 key GDPR topics. Therefore, K-means clustering based summariser is beaten by the PDC-based model in terms of SSD to GDPR topic vectors.

\begin{table}[]
\centering
\begin{tabular}{|c|r|r|r|r|}
\hline
\textbf{Company} & \multicolumn{1}{c|}{\textbf{Random}} & \multicolumn{1}{c|}{\textbf{PDC}} & \multicolumn{1}{c|}{\textbf{K-means}} & \multicolumn{1}{c|}{\textbf{GDPR (fixed)}} \\ \hline
Accuweather      & 259.28                               & 117.72                            & 154.00                                & 115.08                                     \\ \hline
Adobe            & 271.43                               & 118.39                            & 150.68                                & 115.08                                     \\ \hline
Amazon           & 272.10                               & 127.35                            & 142.79                                & 115.08                                     \\ \hline
...              & \multicolumn{1}{c|}{...}             & \multicolumn{1}{c|}{...}          & \multicolumn{1}{c|}{...}              & \multicolumn{1}{c|}{...}                   \\ \hline
Wordpress        & 266.32                               & 121.32                            & 139.10                                & 115.08                                     \\ \hline
Yahoo            & 272.54                               & 118.71                            & 167.82                                & 115.08                                     \\ \hline
Zoom             & 261.91                               & 106.05                            & 159.13                                & 115.08                                     \\ \hline
\textbf{Mean}    & \textbf{264.52}                      & \textbf{119.68}                   & \textbf{152.06}                       & \textbf{115.08}                            \\ \hline
\textbf{Std. Dev.} & \textbf{6.26}                        & \textbf{12.51}                    & \textbf{14.28}                        & \multicolumn{1}{c|}{\textbf{-}}             \\ \hline
\end{tabular}
\caption{SSD evaluation results of each company's PP document, including the mean and standard deviation.}
\label{tab:ssdresults}
\end{table}

\subsection{ROUGE}
\begin{figure}[h!]
\vskip 5mm
\includegraphics[width=\textwidth]{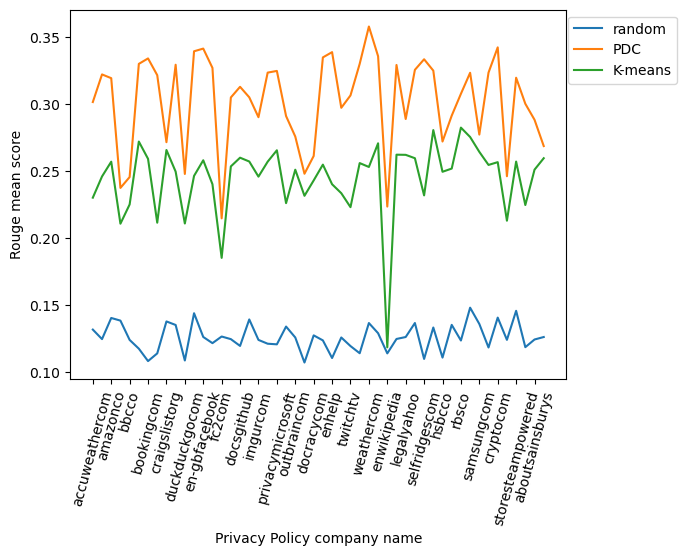}
\centering
    \caption{A visual representation of the ROUGE score evaluation results. Only half of the company names are displayed on the X-axis for better readability. The ROUGE score used for this figure is calculated by averaging the ROUGE-1 score and ROUGE-L score.}
\label{fig:rouge_plot}
\end{figure}

Table \ref{tab:rougeresults} shows the ROUGE-1 and ROUGE-L f-scores of the summaries generated by the PDC-clustering based summariser and the K-means clustering based summariser, along with the scores computed for the randomly generated sentences mostly irrelevant to the topic (the random baseline model for benchmarking). The PDC-clustering based summariser recorded f-scores of 0.3095 (ROUGE-1) and 0.3073 (ROUGE-L), while the K-means model resulted in 0.2491 (ROUGE-1) and 0.2493 (ROUGE-L) 0.2448. The difference in the ROUGE scores of the PDC-clustering based summariser and the K-means summariser remained consistent to be around 23\% - 24\%, therefore it could be claimed that the PDC-based summariser outperformed its competitor by some margin. Also, the ROUGE-1 scores and the ROUGE-L scores had only a marginal difference and analysing the ROUGE score in two separate metrics did not create meaningful results. Hence, it is decided to use the average of the ROUGE-1 score and the ROUGE-L score from further on to represent the overall ROUGE score. Figure \ref{fig:rouge_plot} portrays the mean ROUGE scores of the generated summaries and random sentences, by each company's privacy policy. The figure (\ref{fig:rouge_plot}) resembles figure \ref{fig:ssd_plot} in that the score lines representing the PDC-based summariser and K-means based summariser share a tendency to move in a similar direction. It is also repeated with the mean Rouge scores that the K-means based summariser is outperformed by the PDC-based summariser on all the PP documents. With Wikipedia's privacy policy document, the K-means based summariser almost drew to a random collection of sentences. The values for GDPR sample sentences are disregarded in this figure (\ref{fig:rouge_plot}) in contrast to figure \ref{fig:ssd_plot}, as the sentences are used as the gold standard.

Due to limited human resources allocated for this work, having website-specific human-annotated summaries was difficult to be chosen as an option to perform ROUGE evaluation on the generated summaries. Instead, the combined GDPR-suggested sample sentences were used as the gold standard. This is a limitation and would be discussed further in a dedicated section. Therefore, there could be generated sentences that have similar meanings to the gold standard but the specific use of vocabularies and terminologies are somewhat different. This could cause a significant limitation to the score because the Rouge score utilises exact matches of words for evaluation regardless of specific models N, L or W. To overcome this limitation to some extent, Rouge-1 and Rouge-L which have relatively more flexibility are used as the main models of evaluation. Accordingly, the ROUGE score that appears in table \ref{tab:rougeresults} and figure \ref{fig:rouge_plot} is acquired by averaging the ROUGE-1 score and the ROUGE-L score. This matter would be further discussed in the \textit{Limitations} section in chapter 5. 

\begin{table}[]
\centering
\begin{tabular}{|c|ll|ll|ll|}
\hline
\multirow{2}{*}{\textbf{Company}} & \multicolumn{2}{c|}{\textbf{Random}}                                          & \multicolumn{2}{c|}{\textbf{PDC}}                                             & \multicolumn{2}{c|}{\textbf{K-means}}                                         \\ \cline{2-7} 
                                  & \multicolumn{1}{c|}{\textbf{R-1}} & \multicolumn{1}{c|}{\textbf{R-L}} & \multicolumn{1}{c|}{\textbf{R-1}} & \multicolumn{1}{c|}{\textbf{R-L}} & \multicolumn{1}{c|}{\textbf{R-1}} & \multicolumn{1}{c|}{\textbf{R-L}} \\ \hline
Accuweather                       & \multicolumn{1}{l|}{0.0840}           & 0.1126                                & \multicolumn{1}{l|}{0.3204}           & 0.3037                                & \multicolumn{1}{l|}{0.2313}           & 0.2350                                \\ \hline
Adobe                             & \multicolumn{1}{l|}{0.1258}           & 0.1504                                & \multicolumn{1}{l|}{0.3257}           & 0.3256                                & \multicolumn{1}{l|}{0.2451}           & 0.2513                                \\ \hline
Amazon                            & \multicolumn{1}{l|}{0.1354}           & 0.1581                                & \multicolumn{1}{l|}{0.3332}           & 0.3129                                & \multicolumn{1}{l|}{0.2565}           & 0.2658                                \\ \hline
...                               & \multicolumn{1}{c|}{...}              & \multicolumn{1}{c|}{...}              & \multicolumn{1}{c|}{...}              & \multicolumn{1}{c|}{...}              & \multicolumn{1}{c|}{...}              & \multicolumn{1}{c|}{...}              \\ \hline
Wordpress                         & \multicolumn{1}{l|}{0.1276}           & 0.1502                                & \multicolumn{1}{l|}{0.3261}           & 0.3416                                & \multicolumn{1}{l|}{0.2736}           & 0.2578                                \\ \hline
Yahoo                             & \multicolumn{1}{l|}{0.0957}           & 0.1242                                & \multicolumn{1}{l|}{0.3142}           & 0.2916                                & \multicolumn{1}{l|}{0.2771}           & 0.2616                                \\ \hline
Zoom                              & \multicolumn{1}{l|}{0.1080}           & 0.1396                                & \multicolumn{1}{l|}{0.3476}           & 0.3113                                & \multicolumn{1}{l|}{0.2692}           & 0.2601                                \\ \hline
\textbf{Mean}                     & \multicolumn{1}{l|}{\textbf{0.1163}}  & \textbf{0.1405}                       & \multicolumn{1}{l|}{\textbf{0.3095}}  & \textbf{0.3073}                       & \multicolumn{1}{l|}{\textbf{0.2491}}  & \textbf{0.2493}                       \\ \hline
Std. Dev.                         & \multicolumn{1}{l|}{0.0128}           & 0.0113                                & \multicolumn{1}{l|}{0.0401}           & 0.0304                                & \multicolumn{1}{l|}{0.0308}           & 0.0250                                \\ \hline
\end{tabular}
\caption{Rouge score evaluation results of each company's PP document, including the mean and standard deviation. As explained in the last part of the section, only ROUGE-1 and ROUGE-L scores are used, therefore ROUGE-W, ROUGE-2 and ROUGE-3 results are disregarded. R-1 is an abbreviation of ROUGE-1, and R-L is short for ROUGE-L.}
\label{tab:rougeresults}
\end{table}

\chapter{Conclusions}

\section{Conclusion}
This work aims to facilitate the readings of Privacy Policy (PP) documents by automatically summarising the documents by summarisation mechanisms based on clustering algorithms. The summarisation tool implemented in this work is based on an extractive summarisation system as introduced in section 2.7, rather than a human-like summariser that follows a read-understand-rewrite procedure (or abstractive summarisation). Two clustering algorithms (Pre-determined centroid (PDC) clustering and K-means clustering) are used to implement the summarisers, and two evaluation methods (Sum of Squared Distance (SSD) and ROUGE \cite{lin2004rouge} score) are used to figure the usefulness of the summarisation systems. Sentence vectorisation of PP document sentences using SBERT \cite{reimers2019sentencebert} package was done for natural language encoding. These sentence vectors are used to calculate distances between each of them and pre-determined potential cluster centres. By focusing on the essential 14 topics suggested in GDPR \cite{wodford_2019}, both summarisers target to find sentences in a PP document that have similar meanings to the topics. An example of the generated summaries is displayed in the Appendix \ref{tab:examplesummary}. The SSD evaluation calculates the sum of squared distances between vectorised topics and each selected sentence to find the lowest sum. ROUGE evaluation considers exact matches of N-grams, longest sequences and weighted longest sequences, depending on the choice of the model options. For both evaluation methods, the PDC-clustering based summariser outperformed its K-means clustering based counterpart by 27\% and 24\% respectively. This is believed to be attributed to the fact that the PDC-clustering based model is trained in a semi-supervised way based on pre-defined cluster centres which are the 14 essential GDPR topics, while the K-means model is trained on a list of general PP documents in an unsupervised manner.

It is demonstrated in this work that performing task-specific tweaks and fine-tunes on algorithms and datasets for machine-learning tasks is effective. Considering that K-means clustering can be done in a completely unsupervised fashion whereas PDC-clustering requires human efforts to provide centroid data, it would be generally a matter of making a trade-off between accuracy and feasibility. However, it would be recommended to use the PDC-clustering based summariser instead of K-means for the specific task introduced in this paper, and this is not a matter of making a trade-off. This is because: First, it is easy to predefine the centroids for this task, knowing that the 14 essential GDPR topics \cite{wodford_2019} are specified by the EU and publicly accessible. Second, as has been explained by the evaluation scores, the PDC-clustering based model is better at finding sentences close to the topic vectors than the K-means model.

A practical use case of the summariser models implemented in this work could be testing the GDPR compliance of a PP document. Due to its extractive summarisation design, it could specifically look for sentences related to the 14 essential GDPR topics. Once the summariser spots the best candidate sentences for each topic, SSD can be used as an evaluation metric to test if all the essential topics of GDPR are covered in the PP document. If the SSD score for a certain topic is lower than a threshold value, the company that published the document could be ordered to rewrite that part to meet the requirements and potentially be penalised if it fails to do so.

\section{Limitations}

As briefly explained in the previous chapter, the ROUGE score only focuses on exact matches of words or characters. This means that words with similar meanings but in different representations are ignored or often negatively affect the scores. For instance, the two sentences \textit{"We retain your personal data until you unregister from our website"} and \textit{"Your data is stored safely in our database and will be removed when you delete your account"} only have three matching words even though they basically mean the same thing. This aspect of the evaluation metric became a limitation for correctly evaluating the effectiveness of this work. Therefore, the average F-scores for the PDC-clustering based summariser and the K-means clustering based summariser were limited to be in the range of 0.31 and 0.24, signalling that the summaries might not have many of the matching sequences or n-grams. This may be because both of the summarisation mechanisms were designed and implemented using sentence transformers \cite{reimers2019sentencebert}, whose focus is on encoding the semantic meaning differences/similarities between sentences, instead of finding matching sequences. Accordingly, this limitation does not appear on the SSD evaluation because it calculates the Euclidean distances between sentence vectors encoded to represent semantic meanings, instead of looking for exact matches. In conclusion, SSD can be a better way compared to ROUGE to evaluate how close the summary sentences found by the summarisation models are to the topic vectors, where "close" indicates the similarity in the semantic meanings between sentences.

Also in this work, the GDPR sample sentence "compounds" are used as the gold standard summaries in the ROUGE summary evaluation mechanism due to the difficulty of generating human-written sentences for each of the 50 PP documents given the limited time budget. This means that the general sample sentences that may lack specific details are used without fine-tuning to represent different PP documents from various websites (even though they may include the essential topics suggested in GDPR). Therefore, the evaluation may not be completely accurate compared to performing the ROUGE evaluation on human-written summaries targeted to explain each PP document individually.

Another limiting aspect of this work would be that the selected website pages that include PP document texts often had dedicated pages for different sections, such as a cookies policy page or a data policy page, etc. Tracking all the page URLs for separate sections were foreseen to be a labouriously demanding job, therefore it could not be done in this work. The exact URLs that only include the main text for PP documents are used, therefore the summarisers (regardless of the models) might have been confused while searching for sentences most relevant to cluster centres as they will never find the global best search results from that page (because they do not exist in the page regardless of the performance of the model). This may have been another likely factor that reduced the performance of the models.

\section{Future work}

Potential future works to be done would include overcoming the limitations explained in the previous section. First, assuming that there would be some financial resources allowed for the project, human-written summaries of the PP documents could be earned by utilising crowd-sourcing services such as Amazon Mechanical Turk \cite{crowston2012amazon}. Additionally, by using the service, PP document-related texts that are located outside of the main PP document page such as the cookies policy page and the data policy page could be acquired as well. This way, the gold standard summary would be more accurate in representing the individual PP documents, and the summarisers could perform to their full potential. Therefore, both the SSD and Rouge evaluation mechanisms could be more accurate in measuring the performances of the summariser models.

Also, the functionality of the summariser could be extended to not only searching for the sentences within the boundaries of the essential topics but to classify them by subtopics. For instance, some PP documents distributed the information on the types of data they collect from the user in several separate sentences. That is, instead of putting \textit{"We collect A, B and C"}, some websites put \textit{"We collect:"} followed by three separate lines \textit{"A", "B"} and \textit{"C"}. In these cases, when the value for the parameter \textit{n\_best} is small (\textit{e.g.} 3 or lower) then only a chunk of the data types they collect are retrieved by the summariser (the summariser may return \textit{"We collect:"} and \textit{"B"}, using the example above). In future, I would like to build a module that collects all sentences relevant to the topic (can be limited to targeted topics such as data types collected or types of cookies used, etc.) and merge them into a single sentence to be easily human-readable. Once implemented, this extension is expected to mitigate the problem of returning incomplete information for certain topics.

Another future work that could be done is designing and implementing a web-based front-end application to test out the usability of the summariser model. The web application would show the summaries generated by each of the two summarisation models and allow the user to switch between the two models so that he can compare the results and decide which one to read through. Once the front-end implementation is completed, a user-survey targeting 100 or more participants would be hosted where the questions are mostly focused on the summary performance of the chosen model.

\bibliographystyle{plain}
\bibliography{references.bib}

\begin{thebibliography}{10}

\bibitem{wodford_2019}
Ben wolford. writing a gdpr-compliant privacy notice (template included). https://gdpr.eu/privacy-notice/, Feb 2019.

\bibitem{gdpr_2019}
General data protection regulation. https://gdpr-info.eu/, Sep 2019.

\bibitem{randomwordgenerator2022}
random sentence generator. https://randomwordgenerator.com/sentence.php, 2022.

\bibitem{ankerst1999optics}
Mihael Ankerst, Markus~M Breunig, Hans-Peter Kriegel, and J{\"o}rg Sander.
\newblock Optics: Ordering points to identify the clustering structure.
\newblock {\em ACM Sigmod record}, 28(2):49--60, 1999.

\bibitem{pyrouge}
Diego Antognini.
\newblock Py-rouge.
\newblock https://pypi.org/project/py-rouge/\#description, 2018.

\bibitem{armstrong_2021}
Martin Armstrong.
\newblock Infographic: How many websites are there? https://www.statista.com/chart/19058/number-of-websites-online/, Aug 2021.

\bibitem{bair2013semi}
Eric Bair.
\newblock Semi-supervised clustering methods.
\newblock {\em Wiley Interdisciplinary Reviews: Computational Statistics}, 5(5):349--361, 2013.

\bibitem{bannihatti2020finding}
Vinayshekhar Bannihatti~Kumar, Roger Iyengar, Namita Nisal, Yuanyuan Feng, Hana Habib, Peter Story, Sushain Cherivirala, Margaret Hagan, Lorrie Cranor, Shomir Wilson, et~al.
\newblock Finding a choice in a haystack: Automatic extraction of opt-out statements from privacy policy text.
\newblock In {\em Proceedings of The Web Conference 2020}, pages 1943--1954, 2020.

\bibitem{basu2002semi}
Sugato Basu, Arindam Banerjee, and Raymond Mooney.
\newblock Semi-supervised clustering by seeding.
\newblock In {\em In Proceedings of 19th International Conference on Machine Learning (ICML-2002}. Citeseer, 2002.

\bibitem{bergmann2007automatic}
Mike Bergmann, Simone Fischer-H{\"u}bner, Andreas Pfitzmann, Marit Hansen, and John~S{\"o}ren Pettersson.
\newblock Automatic privacy policy clustering.
\newblock {\em Presentation at IFIP Third International Summer School “The Future of Identity in the Information Society”, Karlstad University, Sweden, 6th-10th August}, 2007.

\bibitem{bowman2015large}
Samuel~R Bowman, Gabor Angeli, Christopher Potts, and Christopher~D Manning.
\newblock A large annotated corpus for learning natural language inference.
\newblock {\em arXiv preprint arXiv:1508.05326}, 2015.

\bibitem{celikyilmaz2018deep}
Asli Celikyilmaz, Antoine Bosselut, Xiaodong He, and Yejin Choi.
\newblock Deep communicating agents for abstractive summarization.
\newblock {\em arXiv preprint arXiv:1803.10357}, 2018.

\bibitem{cherivirala2016visualization}
SK~Cherivirala, F~Schaub, MS~Andersen, S~Wilson, N~Sadeh, and JR~Reidenberg.
\newblock Visualization and interactive exploration of data practices in privacy policies.
\newblock In {\em Symposium on Usable Privacy and Security}, pages 3--10, 2016.

\bibitem{choi2020worth}
Chanwoo Choi.
\newblock {\em Worth-reading finder: A machine-learnt solution to find ”important” clauses in Privacy Policy documents based on 3 different personas}.
\newblock {Bachelor's} dissertation, The University of Edinburgh, 2020.

\bibitem{crowston2012amazon}
Kevin Crowston.
\newblock Amazon mechanical turk: A research tool for organizations and information systems scholars.
\newblock In {\em Shaping the future of ict research. methods and approaches}, pages 210--221. Springer, 2012.

\bibitem{derpanis2005mean}
Konstantinos~G Derpanis.
\newblock Mean shift clustering.
\newblock {\em Lecture Notes}, 32, 2005.

\bibitem{devlin2018bert}
Jacob Devlin, Ming-Wei Chang, Kenton Lee, and Kristina Toutanova.
\newblock Bert: Pre-training of deep bidirectional transformers for language understanding.
\newblock {\em arXiv preprint arXiv:1810.04805}, 2018.

\bibitem{dueck2009affinity}
Delbert Dueck.
\newblock {\em Affinity propagation: clustering data by passing messages}.
\newblock Citeseer, 2009.

\bibitem{eick2004supervised}
Christoph~F Eick, Nidal Zeidat, and Zhenghong Zhao.
\newblock Supervised clustering-algorithms and benefits.
\newblock In {\em 16Th IEEE international conference on tools with artificial intelligence}, pages 774--776. IEEE, 2004.

\bibitem{gupta2012multi}
Virendra~Kumar Gupta and Tanveer~J Siddiqui.
\newblock Multi-document summarization using sentence clustering.
\newblock In {\em 2012 4th International Conference on Intelligent Human Computer Interaction (IHCI)}, pages 1--5. IEEE, 2012.

\bibitem{habib2020s}
Hana Habib, Sarah Pearman, Jiamin Wang, Yixin Zou, Alessandro Acquisti, Lorrie~Faith Cranor, Norman Sadeh, and Florian Schaub.
\newblock " it's a scavenger hunt": Usability of websites' opt-out and data deletion choices.
\newblock In {\em Proceedings of the 2020 CHI Conference on Human Factors in Computing Systems}, pages 1--12, 2020.

\bibitem{hahsler2019dbscan}
Michael Hahsler, Matthew Piekenbrock, and Derek Doran.
\newblock dbscan: Fast density-based clustering with r.
\newblock {\em Journal of Statistical Software}, 91:1--30, 2019.

\bibitem{hochreiter1997long}
Sepp Hochreiter and J{\"u}rgen Schmidhuber.
\newblock Long short-term memory.
\newblock {\em Neural computation}, 9(8):1735--1780, 1997.

\bibitem{hong2020selecting}
David Hong, Yue Sheng, and Edgar Dobriban.
\newblock Selecting the number of components in pca via random signflips.
\newblock {\em arXiv preprint arXiv:2012.02985}, 2020.

\bibitem{jain2010data}
Anil~K Jain.
\newblock Data clustering: 50 years beyond k-means.
\newblock {\em Pattern recognition letters}, 31(8):651--666, 2010.

\bibitem{krause1973taxicab}
Eugene~F Krause.
\newblock Taxicab geometry.
\newblock {\em The Mathematics Teacher}, 66(8):695--706, 1973.

\bibitem{lin2004rouge}
Chin-Yew Lin.
\newblock Rouge: A package for automatic evaluation of summaries.
\newblock In {\em Text summarization branches out}, pages 74--81, 2004.

\bibitem{liu2018towards}
Frederick Liu, Shomir Wilson, Peter Story, Sebastian Zimmeck, and Norman Sadeh.
\newblock Towards automatic classification of privacy policy text.
\newblock {\em School of Computer Science Carnegie Mellon University, Pittsburgh, PA, Tech. Rep. CMU-ISR-17-118R and CMULTI-17-010}, 2018.

\bibitem{liu2019text}
Yang Liu and Mirella Lapata.
\newblock Text summarization with pretrained encoders.
\newblock {\em arXiv preprint arXiv:1908.08345}, 2019.

\bibitem{lu1978sentence}
Shin-Yee Lu and King~Sun Fu.
\newblock A sentence-to-sentence clustering procedure for pattern analysis.
\newblock {\em IEEE Transactions on Systems, Man, and Cybernetics}, 8(5):381--389, 1978.

\bibitem{macqueen1967some}
James MacQueen et~al.
\newblock Some methods for classification and analysis of multivariate observations.
\newblock In {\em Proceedings of the fifth Berkeley symposium on mathematical statistics and probability}, volume~1, pages 281--297. Oakland, CA, USA, 1967.

\bibitem{mcdonald2008cost}
Aleecia~M McDonald and Lorrie~Faith Cranor.
\newblock The cost of reading privacy policies.
\newblock {\em Isjlp}, 4:543, 2008.

\bibitem{mcinnes2017hdbscan}
Leland McInnes, John Healy, and Steve Astels.
\newblock hdbscan: Hierarchical density based clustering.
\newblock {\em J. Open Source Softw.}, 2(11):205, 2017.

\bibitem{mikolov2010recurrent}
Tomas Mikolov, Martin Karafi{\'a}t, Lukas Burget, Jan Cernock{\`y}, and Sanjeev Khudanpur.
\newblock Recurrent neural network based language model.
\newblock In {\em Interspeech}, volume~2, pages 1045--1048. Makuhari, 2010.

\bibitem{moradi2018cibs}
Milad Moradi.
\newblock Cibs: A biomedical text summarizer using topic-based sentence clustering.
\newblock {\em Journal of biomedical informatics}, 88:53--61, 2018.

\bibitem{mullner2011modern}
Daniel M{\"u}llner.
\newblock Modern hierarchical, agglomerative clustering algorithms.
\newblock {\em arXiv preprint arXiv:1109.2378}, 2011.

\bibitem{narayan2018don}
Shashi Narayan, Shay~B Cohen, and Mirella Lapata.
\newblock Don't give me the details, just the summary! topic-aware convolutional neural networks for extreme summarization.
\newblock {\em arXiv preprint arXiv:1808.08745}, 2018.

\bibitem{obar2020biggest}
Jonathan~A Obar and Anne Oeldorf-Hirsch.
\newblock The biggest lie on the internet: Ignoring the privacy policies and terms of service policies of social networking services.
\newblock {\em Information, Communication \& Society}, 23(1):128--147, 2020.

\bibitem{pearson1901liii}
Karl Pearson.
\newblock Liii. on lines and planes of closest fit to systems of points in space.
\newblock {\em The London, Edinburgh, and Dublin philosophical magazine and journal of science}, 2(11):559--572, 1901.

\bibitem{scikitLearn}
F.~Pedregosa, G.~Varoquaux, A.~Gramfort, V.~Michel, B.~Thirion, O.~Grisel, M.~Blondel, P.~Prettenhofer, R.~Weiss, V.~Dubourg, J.~Vanderplas, A.~Passos, D.~Cournapeau, M.~Brucher, M.~Perrot, and E.~Duchesnay.
\newblock Scikit-learn: Machine learning in {P}ython.
\newblock {\em Journal of Machine Learning Research}, 12:2825--2830, 2011.

\bibitem{qin1998determining}
S~Joe Qin and Ricardo Dunia.
\newblock Determining the number of principal components for best reconstruction.
\newblock {\em IFAC Proceedings Volumes}, 31(11):357--362, 1998.

\bibitem{qiu2020pre}
Xipeng Qiu, Tianxiang Sun, Yige Xu, Yunfan Shao, Ning Dai, and Xuanjing Huang.
\newblock Pre-trained models for natural language processing: A survey.
\newblock {\em Science China Technological Sciences}, 63(10):1872--1897, 2020.

\bibitem{radford2019language}
Alec Radford, Jeffrey Wu, Rewon Child, David Luan, Dario Amodei, Ilya Sutskever, et~al.
\newblock Language models are unsupervised multitask learners.
\newblock {\em OpenAI blog}, 1(8):9, 2019.

\bibitem{rajpurkar2016squad}
Pranav Rajpurkar, Jian Zhang, Konstantin Lopyrev, and Percy Liang.
\newblock Squad: 100,000+ questions for machine comprehension of text.
\newblock {\em arXiv preprint arXiv:1606.05250}, 2016.

\bibitem{reimers2019sentencebert}
Nils Reimers and Iryna Gurevych.
\newblock Sentence-bert: Sentence embeddings using siamese bert-networks.
\newblock In {\em Proceedings of the 2019 Conference on Empirical Methods in Natural Language Processing}. Association for Computational Linguistics, 11 2019.

\bibitem{rousseeuw1987Silhouettes}
Peter~J Rousseeuw.
\newblock Silhouettes: a graphical aid to the interpretation and validation of cluster analysis.
\newblock {\em Journal of computational and applied mathematics}, 20:53--65, 1987.

\bibitem{sculley2010web}
David Sculley.
\newblock Web-scale k-means clustering.
\newblock In {\em Proceedings of the 19th international conference on World wide web}, pages 1177--1178, 2010.

\bibitem{steinfeld2016agree}
Nili Steinfeld.
\newblock “i agree to the terms and conditions”:(how) do users read privacy policies online? an eye-tracking experiment.
\newblock {\em Computers in human behavior}, 55:992--1000, 2016.

\bibitem{sun2019fine}
Chi Sun, Xipeng Qiu, Yige Xu, and Xuanjing Huang.
\newblock How to fine-tune bert for text classification?
\newblock In {\em China national conference on Chinese computational linguistics}, pages 194--206. Springer, 2019.

\bibitem{vaswani2017attention}
Ashish Vaswani, Noam Shazeer, Niki Parmar, Jakob Uszkoreit, Llion Jones, Aidan~N Gomez, {\L}ukasz Kaiser, and Illia Polosukhin.
\newblock Attention is all you need.
\newblock In {\em Advances in neural information processing systems}, pages 5998--6008, 2017.

\bibitem{voigt2017eu}
Paul Voigt and Axel Von~dem Bussche.
\newblock The eu general data protection regulation (gdpr).
\newblock {\em A Practical Guide, 1st Ed., Cham: Springer International Publishing}, 10(3152676):10--5555, 2017.

\bibitem{wang2004image}
Jue Wang, Bo~Thiesson, Yingqing Xu, and Michael Cohen.
\newblock Image and video segmentation by anisotropic kernel mean shift.
\newblock In {\em European conference on computer vision}, pages 238--249. Springer, 2004.

\bibitem{westin2019opt}
Fiona Westin and Sonia Chiasson.
\newblock Opt out of privacy or" go home" understanding reluctant privacy behaviours through the fomo-centric design paradigm.
\newblock In {\em Proceedings of the New Security Paradigms Workshop}, pages 57--67, 2019.

\bibitem{williams2017broad}
Adina Williams, Nikita Nangia, and Samuel~R Bowman.
\newblock A broad-coverage challenge corpus for sentence understanding through inference.
\newblock {\em arXiv preprint arXiv:1704.05426}, 2017.

\bibitem{yang2014enhancing}
Libin Yang, Xiaoyan Cai, Yang Zhang, and Peng Shi.
\newblock Enhancing sentence-level clustering with ranking-based clustering framework for theme-based summarization.
\newblock {\em Information sciences}, 260:37--50, 2014.

\bibitem{yang2012robust}
Miin-Shen Yang, Chien-Yo Lai, and Chih-Ying Lin.
\newblock A robust em clustering algorithm for gaussian mixture models.
\newblock {\em Pattern Recognition}, 45(11):3950--3961, 2012.

\bibitem{zhang1996birch}
Tian Zhang, Raghu Ramakrishnan, and Miron Livny.
\newblock Birch: an efficient data clustering method for very large databases.
\newblock {\em ACM sigmod record}, 25(2):103--114, 1996.

\bibitem{zhang2018neural}
Xingxing Zhang, Mirella Lapata, Furu Wei, and Ming Zhou.
\newblock Neural latent extractive document summarization.
\newblock {\em arXiv preprint arXiv:1808.07187}, 2018.

\bibitem{zhang2010understanding}
Yin Zhang, Rong Jin, and Zhi-Hua Zhou.
\newblock Understanding bag-of-words model: a statistical framework.
\newblock {\em International Journal of Machine Learning and Cybernetics}, 1(1):43--52, 2010.

\bibitem{zhou2018neural}
Qingyu Zhou, Nan Yang, Furu Wei, Shaohan Huang, Ming Zhou, and Tiejun Zhao.
\newblock Neural document summarization by jointly learning to score and select sentences.
\newblock {\em arXiv preprint arXiv:1807.02305}, 2018.

\bibitem{zhu2020incorporating}
Jinhua Zhu, Yingce Xia, Lijun Wu, Di~He, Tao Qin, Wengang Zhou, Houqiang Li, and Tie-Yan Liu.
\newblock Incorporating bert into neural machine translation.
\newblock {\em arXiv preprint arXiv:2002.06823}, 2020.

\bibitem{zhu2019classification}
Qiuyu Zhu and Ruixin Zhang.
\newblock A classification supervised auto-encoder based on predefined evenly-distributed class centroids.
\newblock {\em arXiv preprint arXiv:1902.00220}, 2019.

\end{thebibliography}

\appendix
\renewcommand\chaptername{Appendix}
\chapter{Evaluation results (SSD)} 
\begin{table}[]
\centering
\small
\begin{tabular}{|c|c|c|c|c|c|}
\hline
\textbf{No.} & \multicolumn{1}{c|}{\textbf{Company}} &  \multicolumn{1}{c|}{\textbf{Random}} & \multicolumn{1}{c|}{\textbf{PDC}} & \multicolumn{1}{c|}{\textbf{K-means}} & \multicolumn{1}{c|}{\textbf{GDPR (fixed)}} \\ \hline
01 & Accuweather         &259.2791 &117.7154 &153.9983 & 115.078                    \\ \hline
02 &Adobe               & 271.4259 & 118.3919 & 150.6797 &  115.078                    \\ \hline
03 &Amazon              & 272.0955 & 127.3489 & 142.7929 &  115.078                    \\ \hline
04 &Apple               &  269.6002 & 133.4980 & 170.9003 &   115.078                    \\ \hline
05 &BBC                 &  257.3826 & 130.5527 & 149.4780 &   115.078                    \\ \hline
06 &Bit.ly              &  250.5208 & 115.6109 & 149.0089 &   115.078                    \\ \hline
07 &Booking.com         & 261.8183 & 112.8720 & 137.8312 &   115.078                    \\ \hline
08 &CNN                 &  265.4834 & 115.4481 & 156.3346 &   115.078                    \\ \hline
09 &Craigslist          &  254.1572 & 135.6693 & 168.7298 &   115.078                    \\ \hline
10 &Discord             &  265.2553 & 105.1960 & 140.9012 &   115.078                    \\ \hline
11 &DuckDuckgo          &  264.4286 & 127.8933 & 165.5231 &   115.078                    \\ \hline
12 &Ebay                &  260.0103 & 118.7229 & 154.7026 &   115.078                    \\ \hline
13 &Facebook            &  258.8454 & 129.5140 & 161.2143 &   115.078                    \\ \hline
14 &Fandom              &  269.3330 & 116.9396 & 146.5154 &   115.078                    \\ \hline
15 &fc2                 &  264.3741 & 146.7468 & 164.9264 &   115.078                    \\ \hline
16 &Fox news            &  266.2818 & 129.3519 & 147.5297 &   115.078                    \\ \hline
17 &Github              &  265.1300 & 124.1395 & 151.0889 &   115.078                    \\ \hline
18 &Imdb                &  265.7324 & 120.3096 & 153.5037 &   115.078                    \\ \hline
19 &imgur               &  269.1370 & 131.6383 & 154.7619 &   115.078                    \\ \hline
20 &Linkedin            &  260.0946 & 105.2546 & 140.5771 &   115.078                    \\ \hline
21 &Microsoft           &  268.1959 & 119.9294 & 146.8715 &   115.078                    \\ \hline
22 &Netflix             &  250.9792 & 117.2731 & 152.1793 &   115.078                    \\ \hline
23 &Outbrain            &  258.3041 & 121.5575 & 132.8060 &   115.078                    \\ \hline
24 &Pinterest           &  275.5505 & 114.6272 & 122.9251 &   115.078                    \\ \hline
25 &Quora               &  273.1365 & 154.3024 & 176.9815 &   115.078                   \\ \hline
26 &Reddit              &  265.8606 & 120.5610 & 163.6190 &   115.078                   \\ \hline
27 &Roblox              &  273.4940 & 111.2047 & 141.2322 &   115.078                   \\ \hline
28 &Tumblr              &  256.8898 & 120.4684 & 165.7294 &   115.078                   \\ \hline
29 &Twitch              &  260.5789 & 134.2332 & 174.3503 &   115.078                   \\ \hline
30 &Twitter             &  270.0506 & 113.3764 & 126.3804 &   115.078                  \\ \hline
31 &weather.com         &  274.9953 & 109.1715 & 149.5505 &   115.078                  \\ \hline
32 &Whatsapp            &  258.0421 & 106.7192 & 150.3537 &   115.078                  \\ \hline
33 &Wikipedia           &  258.9463 & 128.0081 & 152.3977 &   115.078                  \\ \hline
34 &Wordpress           &  266.3189 & 121.3179 & 139.1003 &   115.078                 \\ \hline
35 &Yahoo               &  272.5432 & 118.7098 & 167.8192 &   115.078                 \\ \hline
36 &Zoom                &  261.9080 & 106.0504 & 159.1301 &   115.078                 \\ \hline
37 &Selfridges          &  255.9910 & 104.6784 & 143.7994 &   115.078                \\ \hline
38 &Nike                &  262.7088 & 110.2472 & 130.3867 &   115.078                \\ \hline
39 &HSBC                &  262.7472 & 124.9410 & 162.1418 &   115.078                \\ \hline
40 &Barclays            &  262.6991 & 108.3705 & 161.3803 &   115.078                \\ \hline
41 &RBS.com             &  269.9342 & 111.4676 & 132.9782 &   115.078                 \\ \hline
42 &Deloitte            &  274.1919 & 114.9539 & 140.5276 &   115.078                \\ \hline
43 &Samsung             &  259.5961 & 116.5685 & 163.4381 &   115.078                 \\ \hline
44 &Razer               &  268.2288 & 114.4456 & 166.0414 &   115.078                 \\ \hline
45 &Crypto.com          &  270.4080 & 111.3391 & 139.2075 &   115.078                 \\ \hline
46 &EA                  &  271.2269 & 156.7117 & 173.1594 &   115.078                \\ \hline
47 &Steam               &  259.7036 & 119.6287 & 156.8430 &   115.078                 \\ \hline
48 &Tesco               &  269.9675 &  92.2890 & 189.5367 &   115.078                 \\ \hline
49 &Sainsburys          &  265.8673 &  94.3389 & 134.5320 &   115.078                 \\ \hline
50 &Adidas              &  256.6028 & 123.4880 & 126.5080 &   115.078                 \\ \hline
   &\textbf{Mean}    & \textbf{264.5211}      & \textbf{119.6758}   & \textbf{152.0581}   & \textbf{115.0780}          \\ \hline
   &\textbf{Std. Dev.} & \textbf{6.2588}    & \textbf{12.5138}   & \textbf{14.2830}   & \textbf{-}          \\ \hline
\end{tabular}
\caption{Rouge score evaluation results of each company's PP document, including the mean and standard deviation.}
\end{table}

\chapter{Evaluation results (ROUGE)} 
\begin{table}[]
\centering
\small
\begin{tabular}{|c|c|c|c|c|c|}
\hline
\textbf{No.} & \multicolumn{1}{c|}{\textbf{Company}} & \multicolumn{1}{c|}{\textbf{Random}} & \multicolumn{1}{c|}{\textbf{PDC}} & \multicolumn{1}{c|}{\textbf{K-means}} \\ \hline
01 & Accuweather         & 0.1315   & 0.3015    & 0.2301                    \\ \hline
02 & Adobe               & 0.1244   & 0.3221    & 0.2459                    \\ \hline
03 & Amazon              & 0.1402   & 0.3193    & 0.2569                    \\ \hline
04 & Apple               & 0.1382   & 0.2373    & 0.2106                    \\ \hline
05 & BBC                 & 0.1238   & 0.2455    & 0.2250                    \\ \hline
06 & Bit.ly              & 0.1172   & 0.3300    & 0.2720                    \\ \hline
07 & Booking.com         & 0.1080   & 0.3341    & 0.2591                    \\ \hline
08 & CNN                 & 0.1137   & 0.3216    & 0.2114                    \\ \hline
09 & Craigslist          & 0.1376   & 0.2715    & 0.2657                    \\ \hline
10 & Discord             & 0.1350   & 0.3294    & 0.2495                    \\ \hline
11 & DuckDuckgo          & 0.1084   & 0.2478    & 0.2107                    \\ \hline
12 & Ebay                & 0.1437   & 0.3394    & 0.2463                    \\ \hline
13 & Facebook            & 0.1260   & 0.3414    & 0.2580                    \\ \hline
14 & Fandom              & 0.1214   & 0.3270    & 0.2401                    \\ \hline
15 & fc2                 & 0.1263   & 0.2146    & 0.1851                    \\ \hline
16 & Fox news            & 0.1244   & 0.3049    & 0.2534                    \\ \hline
17 & Github              & 0.1193   & 0.3129    & 0.2599                    \\ \hline
18 & Imdb                & 0.1391   & 0.3050    & 0.2571                    \\ \hline
19 & imgur               & 0.1238   & 0.2902    & 0.2458                    \\ \hline
20 & Linkedin            & 0.1209   & 0.3235    & 0.2569                    \\ \hline
21 & Microsoft           & 0.1205   & 0.3248    & 0.2655                    \\ \hline
22 & Netflix             & 0.1338   & 0.2910    & 0.2260                    \\ \hline
23 & Outbrain            & 0.1256   & 0.2757    & 0.2510                    \\ \hline
24 & Pinterest           & 0.1069   & 0.2479    & 0.2315                    \\ \hline
25 & Quora               & 0.1272   & 0.2612    & 0.2430                    \\ \hline
26 & Reddit              & 0.1234   & 0.3349    & 0.2547                    \\ \hline
27 & Roblox              & 0.1103   & 0.3388    & 0.2401                    \\ \hline
28 & Tumblr              & 0.1256   & 0.2972    & 0.2334                    \\ \hline
29 & Twitch              & 0.1192   & 0.3065    & 0.2230                    \\ \hline
30 & Twitter             & 0.1138   & 0.3300    & 0.2559                    \\ \hline
31 & weather.com         & 0.1364   & 0.3579    & 0.2530                    \\ \hline
32 & Whatsapp            & 0.1288   & 0.3357    & 0.2707                    \\ \hline
33 & Wikipedia           & 0.1137   & 0.2234    & 0.1183                    \\ \hline
34 & Wordpress           & 0.1245   & 0.3292    & 0.2622                    \\ \hline
35 & Yahoo               & 0.1260   & 0.2888    & 0.2621                    \\ \hline
36 & Zoom                & 0.1364   & 0.3256    & 0.2595                    \\ \hline
37 & Selfridges          & 0.1096   & 0.3335    & 0.2318                    \\ \hline
38 & Nike                & 0.1331   & 0.3250    & 0.2806                    \\ \hline
39 & HSBC                & 0.1106   & 0.2720    & 0.2494                    \\ \hline
40 & Barclays            & 0.1350   & 0.2913    & 0.2518                    \\ \hline
41 & RBS.com             & 0.1233   & 0.3077    & 0.2824                    \\ \hline
42 & Deloitte            & 0.1479   & 0.3234    & 0.2754                    \\ \hline
43 & Samsung             & 0.1357   & 0.2772    & 0.2643                    \\ \hline
44 & Razer               & 0.1181   & 0.3255    & 0.2545                    \\ \hline
45 & Crypto.com          & 0.1404   & 0.3424    & 0.2566                    \\ \hline
46 & EA                  & 0.1238   & 0.2461    & 0.2129                    \\ \hline
47 & Steam               & 0.1455   & 0.3197    & 0.2570                    \\ \hline
48 & Tesco               & 0.1184   & 0.3002    & 0.2246                    \\ \hline
49 & Sainsburys          & 0.1241   & 0.2883    & 0.2510                    \\ \hline
50 & Adidas              & 0.1259   & 0.2686    & 0.2595                    \\ \hline
  & \textbf{Mean}    & \textbf{0.1257}      & \textbf{0.3021}   & \textbf{0.2448}             \\ \hline
  & \textbf{Std. Dev.} & \textbf{0.0103}    & \textbf{0.0342}   & \textbf{0.0267}             \\ \hline
\end{tabular}
\caption{Rouge score evaluation results of each company's PP document, including the mean and standard deviation.}
\end{table}

\chapter{GDPR combined sample sentences} 
\begin{table}[]
\begin{tabular}{|l|l|}
\hline
Topic                                      & Sentence Compound                                                                                                                                                                                                                                                                         \\ \hline
what data do we collect?                   & \begin{tabular}[c]{@{}l@{}}we collect personal identification information such as \\ name, email, phone number, etc and other necessary data.\end{tabular}                                                                                                                                \\ \hline
How do we collect your data?               & \begin{tabular}[c]{@{}l@{}}you directly provide us most of the data we collect your \\ data when you register online, place order, voluntarily \\ complete survey, provide feedback, use or view via cookies\end{tabular}                                                                 \\ \hline
How will we use your data?                 & \begin{tabular}[c]{@{}l@{}}we use your data to process order and manage account \\ email you with special offers, share your data with \\ partner companies, and send your data to credit reference \\ agencies to prevent fraud and abuse.\end{tabular}                                  \\ \hline
How do we store your data?                 & \begin{tabular}[c]{@{}l@{}}we securely retain, and maintain your data at until once \\ this period time expired we delete your data by \\ months years.\end{tabular}                                                                                                                      \\ \hline
Marketing                                  & \begin{tabular}[c]{@{}l@{}}we send you information about products and services you \\ might like recommend marketing third party use opt out \\ later right to stop no longer wish marketing purposes.\end{tabular}                                                                       \\ \hline
What are your data protection rights?      & \begin{tabular}[c]{@{}l@{}}your data protection rights you have right to access \\ rectify edit erase remove delete restrict processing \\ object data portable control transfer.\end{tabular}                                                                                            \\ \hline
What are cookies?                          & \begin{tabular}[c]{@{}l@{}}what are cookies cookies are text files placed on your \\ computer when you visit our website we collect \\ through cookies.\end{tabular}                                                                                                                      \\ \hline
How do we use cookies?                     & \begin{tabular}[c]{@{}l@{}}we use your cookies to keep you signed in understand \\ how you use our website.\end{tabular}                                                                                                                                                                  \\ \hline
What types of cookies do we use?           & \begin{tabular}[c]{@{}l@{}}we use different types of cookies, functionality remember \\ your preferences language location advertising links you \\ followed share online data with third parties for advertising \\ authentication security performance analytics research.\end{tabular} \\ \hline
How to manage your cookies                 & \begin{tabular}[c]{@{}l@{}}manage cookies, you can set your browser not to accept \\ cookies remove cookies some of features not function \\ as a result\end{tabular}                                                                                                                     \\ \hline
Privacy policies of other websites         & \begin{tabular}[c]{@{}l@{}}we contain links to other websites our privacy policy apply \\ only to our website, if you click link to another website \\ you should read and refer to their policy'\end{tabular}                                                                            \\ \hline
Changes to our privacy policy              & \begin{tabular}[c]{@{}l@{}}we keep our privacy policy under review and change \\ regularly this was last updated on\end{tabular}                                                                                                                                                          \\ \hline
How to contact us                          & \begin{tabular}[c]{@{}l@{}}how to contact us if you have questions on privacy policy \\ data we hold on you data about data protection rights\end{tabular}                                                                                                                                \\ \hline
How to contact the appropriate authorities & \begin{tabular}[c]{@{}l@{}}how to contact the appropriate authorities and data \\ protection officer report complaint information \\ commissioner office\end{tabular}                                                                                                                     \\ \hline
\end{tabular}
\end{table}

\chapter{Example summary generated by PDC-clustering based summariser}
\begin{table}[]
\centering
\small
\begin{tabular}{|l|}
\hline
\multicolumn{1}{|c|}{{\textbf{Generated Summary}}}                                                                                                                                                                                                                                                                                                                                                                                                                                                                          \\ \hline
\multicolumn{1}{|c|}{\textbf{What data do we collect?}}                                                                                                                                                                                                                                                                                                                                                                                                                                                                         \\ \hline
\begin{tabular}[c]{@{}l@{}}"We also collect contact information if you choose to upload, sync or import it from a device (such as an address \\ book or call log or SMS log history), which we use for things such as helping you and others find people you may\\  know and for the other purposes listed below.."\end{tabular}                                                                                                                                                                                                \\ \hline
"We also use your information to respond to you when you contact us.."                                                                                                                                                                                                                                                                                                                                                                                                                                                          \\ \hline
\multicolumn{1}{|c|}{\textbf{How do we collect your data?}}                                                                                                                                                                                                                                                                                                                                                                                                                                                                     \\ \hline
\begin{tabular}[c]{@{}l@{}}"We collect information about how you use our Products, such as the types of content that you view or engage \\ with, the features you use, the actions you take, the people or accounts you interact with and the time, frequency\\  and duration of your activities."\end{tabular}                                                                                                                                                                                                                 \\ \hline
\begin{tabular}[c]{@{}l@{}}"To create personalised Products that are unique and relevant to you, we use your connections, preferences, \\ interests and activities based on the data that we collect and learn from you and others (including any data \\ with special protections that you choose to provide where you have given your explicit consent); how you \\ use and interact with our Products; and the people, places or things that you're connected to and interested\\  in on and off our Products."\end{tabular} \\ \hline
\multicolumn{1}{|c|}{\textbf{How will we use your data?}}                                                                                                                                                                                                                                                                                                                                                                                                                                                                       \\ \hline
\begin{tabular}[c]{@{}l@{}}"We also receive information about your online and offline actions and purchases from third-party data providers \\ who have the rights to provide us with your information.."\end{tabular}                                                                                                                                                                                                                                                                                                          \\ \hline
\begin{tabular}[c]{@{}l@{}}"Information that we receive about you (including financial transaction data related to purchases made on our \\ Products) can be accessed and preserved for an extended period when it is the subject of a legal request or \\ obligation, governmental investigation or investigations of possible violations of our terms or policies, or \\ otherwise to prevent harm."\end{tabular}                                                                                                             \\ \hline
\multicolumn{1}{|c|}{\textbf{How do we store your data?}}                                                                                                                                                                                                                                                                                                                                                                                                                                                                       \\ \hline
\begin{tabular}[c]{@{}l@{}}"We store data until it is no longer necessary to provide our services and Meta Products, or until your account \\ is deleted – whichever comes first."\end{tabular}                                                                                                                                                                                                                                                                                                                                 \\ \hline
\begin{tabular}[c]{@{}l@{}}"We also retain information from accounts disabled for term breaches for at least a year to prevent repeat \\ abuse or other term breaches.."\end{tabular}                                                                                                                                                                                                                                                                                                                                           \\ \hline
\multicolumn{1}{|c|}{\textbf{How to manage your cookies}}                                                                                                                                                                                                                                                                                                                                                                                                                                                                       \\ \hline
"Learn more about deletion of content that you have shared and cookie data obtained through social plugins.."                                                                                                                                                                                                                                                                                                                                                                                                                   \\ \hline
\begin{tabular}[c]{@{}l@{}}"For example, we will remove developers' access to your Facebook and Instagram data if you haven't used \\ their app in three months, and we are changing login, so that in the next version, we will reduce the data \\ that an app can request without app review to include only name, Instagram username and bio, profile \\ photo and email address."\end{tabular}                                                                                                                              \\ \hline
\multicolumn{1}{|c|}{\textbf{Privacy policies of other websites}}                                                                                                                                                                                                                                                                                                                                                                                                                                                               \\ \hline
\begin{tabular}[c]{@{}l@{}}"We use the information that we have (including your activity off our Products, such as the websites you visit and\\  ads you see) to help advertisers and other partners measure the effectiveness and distribution of their ads and\\  services, and understand the types of people who use their services and how people interact with their \\ websites, apps and services."\end{tabular}                                                                                                        \\ \hline
\begin{tabular}[c]{@{}l@{}}"We provide advertisers with reports about the kinds of people seeing their ads and how their ads are \\ performing, but we don't share information that personally identifies you (information such as your name or \\ email address that by itself can be used to contact you or identifies who you are) unless you give us permission."\end{tabular}                                                                                                                                              \\ \hline
\multicolumn{1}{|c|}{\textbf{Changes to our privacy policy}}                                                                                                                                                                                                                                                                                                                                                                                                                                                                    \\ \hline
\begin{tabular}[c]{@{}l@{}}"You can learn more about cookies and how we use them, and review or change your choice at any time in our \\ Cookie Policy.."\end{tabular}                                                                                                                                                                                                                                                                                                                                                          \\ \hline
\begin{tabular}[c]{@{}l@{}}"We'll notify you before we make changes to this Policy and give you the opportunity to review the revised \\ Policy before you choose to continue using our Products.."\end{tabular}                                                                                                                                                                                                                                                                                                                \\ \hline
\multicolumn{1}{|c|}{\textbf{How to contact us}}                                                                                                                                                                                                                                                                                                                                                                                                                                                                                \\ \hline
"If you have questions about this policy, you can contact us as described below.."                                                                                                                                                                                                                                                                                                                                                                                                                                              \\ \hline
\begin{tabular}[c]{@{}l@{}}"For example, we use data that we have to investigate suspicious activity or violations of our Terms or Policies,\\  or to detect when someone needs help."\end{tabular}                                                                                                                                                                                                                                                                                                                             \\ \hline
\multicolumn{1}{|c|}{\textbf{How to contact the appropriate authorities}}                                                                                                                                                                                                                                                                                                                                                                                                                                                       \\ \hline
\begin{tabular}[c]{@{}l@{}}"You also have the right to lodge a complaint with Meta Platforms Ireland Limited's lead supervisory authority, \\ the Irish Data Protection Commission, or your local supervisory authority.."\end{tabular}                                                                                                                                                                                                                                                                                         \\ \hline
"If you have questions about this policy, you can contact us as described below.."                                                                                                                                                                                                                                                                                                                                                                                                                                              \\ \hline
\end{tabular}
\caption{Example summary generated by the PDC-clustering based model given Facebook's Privacy Policy document at https://en-gb.facebook.com/policy.php.}
\label{tab:examplesummary}
\end{table}

\end{document}